\pdfoutput=1
\documentclass[11pt,letterpaper]{article}

\usepackage[margin=1in]{geometry}
\usepackage{xcolor}
\usepackage[T1]{fontenc}
\usepackage{newtxtext}
\usepackage{newtxmath}
\usepackage{microtype}
\usepackage[bottom]{footmisc}

\usepackage[square,numbers,sort&compress]{natbib}
\usepackage{xurl}
\usepackage{booktabs}
\usepackage{tabularx}
\usepackage{array}
\usepackage{etoolbox}

\usepackage{amsmath}
\usepackage{dsfont}
\newcommand{\mathbbm}[1]{\mathds{#1}}
\usepackage{graphicx}
\usepackage{subcaption}
\usepackage{wrapfig}
\usepackage{algorithm}
\usepackage{algorithmic}
\usepackage{pgfplots}
\pgfplotsset{compat=1.18}
\usepgfplotslibrary{groupplots, fillbetween}
\newcommand{\algname}{QF3}

\usepackage{titlesec}
\titleformat{\section}{\normalfont\Large\bfseries}{\thesection}{0.8em}{}
\titlespacing*{\section}{0pt}{2.2ex plus 0.6ex minus 0.3ex}{0.8ex plus 0.2ex}
\titleformat{\subsection}{\normalfont\large\bfseries}{\thesubsection}{0.8em}{}
\titlespacing*{\subsection}{0pt}{1.8ex plus 0.5ex minus 0.3ex}{0.6ex plus 0.2ex}
\titleformat{\subsubsection}[runin]{\normalfont\normalsize\bfseries}{}{0pt}{}
\titlespacing*{\subsubsection}{0pt}{1.2ex plus 0.4ex minus 0.2ex}{0.5em}
\providecommand{\algname}{QF3}

\definecolor{flopAccentA}{HTML}{2A6F97}
\definecolor{flopAccentB}{HTML}{C98A3B}
\definecolor{flopAccentC}{HTML}{009E73}
\definecolor{flopAccentD}{HTML}{CC79A7}
\definecolor{flopAccentE}{HTML}{6B4F9B}
\definecolor{flopAccentG}{HTML}{8A8A2A}
\definecolor{flopAccentH}{HTML}{CC79A7}
\definecolor{flopAxis}{HTML}{999999}
\definecolor{flopInk}{HTML}{222222}

\pgfplotsset{
  flop curves/.style={
    width=0.165\linewidth,
    height=1.7cm,
    scale only axis,
    axis line style={draw=flopInk, line width=0.5pt},
    tick style={draw=flopInk, line width=0.4pt},
    tick align=outside,
    tick pos=left,
    major tick length=2pt,
    minor tick length=1pt,
    every tick label/.style={font=\sffamily\scriptsize, color=flopInk},
    title style={font=\sffamily\footnotesize, color=flopInk, yshift=-7pt},
    label style={font=\sffamily\scriptsize, color=flopInk},
    every axis x label/.style={
      font=\sffamily\tiny, color=flopInk,
      at={(ticklabel cs:0.5)}, anchor=north, yshift=2pt,
    },
    every axis y label/.style={
      font=\sffamily\tiny, color=flopInk,
      at={(ticklabel cs:0.5)}, anchor=south, rotate=90, xshift=0pt, yshift=-4pt,
    },
    enlarge x limits=0.06,
    enlarge y limits=0.05,
    grid=both,
    major grid style={draw=flopAxis!70, line width=0.3pt, dash pattern=on 1pt off 1.5pt},
    minor grid style={draw=flopAxis!50, line width=0.25pt, dash pattern=on 1pt off 1.5pt},
    minor x tick num=1,
    minor y tick num=1,
    scaled x ticks=false,
    scaled y ticks=false,
    xmin=0, xmax=500000,
    xtick=        {0, 250000, 500000},
    xticklabels=  {0,    250,    500},
    max space between ticks=22pt,
    yticklabel={$\mathsf{\pgfmathprintnumber[fixed, precision=0, 1000 sep={}]{\tick}}$},
    every y tick label/.append style={
      font=\sffamily\scriptsize,
      text width=18pt, align=right, inner xsep=0pt,
    },
    every x tick label/.append style={font=\sffamily\scriptsize},
  },
}

\pgfplotsset{
  flop algo FLOP/.style   ={color=flopAccentA, line width=1.5pt, opacity=1, solid},
  flop algo TD3/.style    ={color=flopAccentB, line width=0.85pt, opacity=0.7, solid},
  flop algo SAC/.style    ={color=flopAccentC, line width=0.85pt, opacity=0.7, solid},
  flop algo QwCFM/.style  ={color=flopAccentD, line width=0.85pt, opacity=0.7, solid},
  flop algo QSM/.style    ={color=flopAccentE, line width=0.85pt, opacity=0.7, solid},
  flop algo FlowRL/.style ={color=flopAccentH, line width=0.85pt, opacity=0.7, solid},
  flop algo DIPO/.style   ={color=flopAccentG, line width=0.85pt, opacity=0.7, solid},
  flop band FLOP/.style   ={fill=flopAccentA, fill opacity=0.18, draw=none},
  flop band TD3/.style    ={fill=flopAccentB, fill opacity=0.035, draw=none},
  flop band SAC/.style    ={fill=flopAccentC, fill opacity=0.035, draw=none},
  flop band QwCFM/.style  ={fill=flopAccentD, fill opacity=0.035, draw=none},
  flop band QSM/.style    ={fill=flopAccentE, fill opacity=0.035, draw=none},
  flop band FlowRL/.style ={fill=flopAccentH, fill opacity=0.035, draw=none},
  flop band DIPO/.style   ={fill=flopAccentG, fill opacity=0.035, draw=none},
}

\expandafter\def\csname flopAlgoLabel@FLOP\endcsname{\algname{}}
\expandafter\def\csname flopAlgoLabel@TD3\endcsname{TD3}
\expandafter\def\csname flopAlgoLabel@SAC\endcsname{SAC}
\expandafter\def\csname flopAlgoLabel@QwCFM\endcsname{Q-weighted CFM}
\expandafter\def\csname flopAlgoLabel@QSM\endcsname{QSM}
\expandafter\def\csname flopAlgoLabel@FlowRL\endcsname{FlowRL}
\expandafter\def\csname flopAlgoLabel@DIPO\endcsname{DIPO}

\expandafter\def\csname flopEnvTitle@HalfCheetah-v4\endcsname{HalfCheetah-v4}
\expandafter\def\csname flopEnvTitle@Hopper-v4\endcsname{Hopper-v4}
\expandafter\def\csname flopEnvTitle@Walker2d-v4\endcsname{Walker2d-v4}
\expandafter\def\csname flopEnvTitle@Ant-v4\endcsname{Ant-v4}
\expandafter\def\csname flopEnvTitle@Humanoid-v4\endcsname{Humanoid-v4}
\newcommand{\flopEnvTitle}[1]{%
  \ifcsname flopEnvTitle@#1\endcsname
    \csname flopEnvTitle@#1\endcsname
  \else
    \detokenize{#1}%
  \fi
}

\newcommand{\flopAxisAnchor}{%
  \addplot[draw=none, forget plot, mark=none] coordinates {(0,0) (500000,1)};
}

\newcommand{\flopAddCurve}[4]{%
  \IfFileExists{#1/#2__#3.csv}{%
    \addplot[name path=flopBand#3lo, draw=none, forget plot]
      table[x=step, y=lo, col sep=comma]{#1/#2__#3.csv};
    \addplot[name path=flopBand#3hi, draw=none, forget plot]
      table[x=step, y=hi, col sep=comma]{#1/#2__#3.csv};
    \addplot[flop band #3, forget plot] fill between[of=flopBand#3lo and flopBand#3hi];
    \ifnum#4=1
      \addplot+[flop algo #3, mark=none]
        table[x=step, y=mean, col sep=comma]{#1/#2__#3.csv};
      \addlegendentry{\csname flopAlgoLabel@#3\endcsname}%
    \else
      \addplot+[flop algo #3, mark=none, forget plot]
        table[x=step, y=mean, col sep=comma]{#1/#2__#3.csv};
    \fi
  }{}%
}

\newcommand{\flopMujocoAllCurvesTwoByTwo}[1]{%
  \begingroup
  \centering
  \begin{tikzpicture}
    \begin{groupplot}[
      group style={
        group size=2 by 2,
        horizontal sep=20pt,
        vertical sep=16pt,
        xticklabels at=edge bottom,
        xlabels at=edge bottom,
      },
      flop curves,
      width=0.34\linewidth,
      height=2.15cm,
      max space between ticks=16pt,
      yticklabel={$\mathsf{\pgfmathparse{\tick/1000}\ifdim\pgfmathresult pt=0pt 0\else\pgfmathprintnumber[fixed, precision=0]{\pgfmathresult}k\fi}$},
      flop mujoco two by two/.try,
      every axis x label/.append style={font=\sffamily\scriptsize},
      every axis y label/.append style={font=\sffamily\scriptsize},
    ]
      \nextgroupplot[
        title={\flopEnvTitle{Hopper-v4}},
        legend to name=flopMujocoLegendAll,
        legend columns=4,
        legend style={
          draw=flopAxis, line width=0.4pt, fill=none,
          font=\sffamily\scriptsize,
          legend cell align=left,
          column sep=8pt,
          inner xsep=4pt, inner ysep=2pt,
          row sep=1pt,
          nodes={font=\sffamily\scriptsize, inner xsep=1pt, inner ysep=1pt, outer ysep=0pt},
        },
        flop mujoco two by two legend/.try,
      ]
      \flopAxisAnchor
      \flopAddCurve{#1}{Hopper-v4}{QSM}{1}%
      \flopAddCurve{#1}{Hopper-v4}{FlowRL}{1}%
      \flopAddCurve{#1}{Hopper-v4}{DIPO}{1}%
      \flopAddCurve{#1}{Hopper-v4}{FLOP}{1}%

      \nextgroupplot[title={\flopEnvTitle{Walker2d-v4}}]
      \flopAxisAnchor
      \flopAddCurve{#1}{Walker2d-v4}{QSM}{0}%
      \flopAddCurve{#1}{Walker2d-v4}{FlowRL}{0}%
      \flopAddCurve{#1}{Walker2d-v4}{DIPO}{0}%
      \flopAddCurve{#1}{Walker2d-v4}{FLOP}{0}%

      \nextgroupplot[title={\flopEnvTitle{Ant-v4}}, xlabel={Env. Steps (K)}]
      \flopAxisAnchor
      \flopAddCurve{#1}{Ant-v4}{QSM}{0}%
      \flopAddCurve{#1}{Ant-v4}{FlowRL}{0}%
      \flopAddCurve{#1}{Ant-v4}{DIPO}{0}%
      \flopAddCurve{#1}{Ant-v4}{FLOP}{0}%

      \nextgroupplot[title={\flopEnvTitle{Humanoid-v4}}, xlabel={Env. Steps (K)}]
      \flopAxisAnchor
      \flopAddCurve{#1}{Humanoid-v4}{QSM}{0}%
      \flopAddCurve{#1}{Humanoid-v4}{FlowRL}{0}%
      \flopAddCurve{#1}{Humanoid-v4}{DIPO}{0}%
      \flopAddCurve{#1}{Humanoid-v4}{FLOP}{0}%
    \end{groupplot}
    \node[rotate=90, anchor=south, font=\sffamily\scriptsize, color=flopInk, yshift=14pt]
      at ($(group c1r1.north west)!0.5!(group c1r2.south west)$) {Eval Return};
  \end{tikzpicture}\par
  \vspace{1pt}
  \pgfplotslegendfromname{flopMujocoLegendAll}%
  \endgroup
}

\pgfplotsset{
  flop clamp curves/.style={
    flop curves,
    width=0.36\linewidth,
    height=3.0cm,
    xmin=0, xmax=250000,
    xtick=        {0, 125000, 250000},
    xticklabels=  {0,    125,    250},
    enlarge x limits=0.04,
  },
}

\pgfplotsset{
  flop algo FQR-clamp/.style    ={color=flopAccentA, line width=1.2pt},
  flop algo FQR-noclamp/.style  ={color=flopAccentB, line width=1.0pt, dash pattern=on 3pt off 2pt},
  flop band  FQR-clamp/.style   ={fill=flopAccentA, fill opacity=0.16, draw=none},
  flop band  FQR-noclamp/.style ={fill=flopAccentB, fill opacity=0.12, draw=none},
}

\expandafter\def\csname flopAlgoLabel@FQR-clamp\endcsname{\algname{}}
\expandafter\def\csname flopAlgoLabel@FQR-noclamp\endcsname{Direct Q (unclamped)}

\newcommand{\flopClampAddCurve}[5]{%
  \IfFileExists{#1/#2__#3.csv}{%
    \addplot[name path=flopClampBand#3lo, draw=none, forget plot]
      table[x=step, y=lo, col sep=comma]{#1/#2__#3.csv};
    \addplot[name path=flopClampBand#3hi, draw=none, forget plot]
      table[x=step, y=hi, col sep=comma]{#1/#2__#3.csv};
    \addplot[flop band #4, forget plot] fill between[of=flopClampBand#3lo and flopClampBand#3hi];
    \ifnum#5=1
      \addplot+[flop algo #4, mark=none]
        table[x=step, y=mean, col sep=comma]{#1/#2__#3.csv};
      \addlegendentry{\csname flopAlgoLabel@#4\endcsname}%
    \else
      \addplot+[flop algo #4, mark=none, forget plot]
        table[x=step, y=mean, col sep=comma]{#1/#2__#3.csv};
    \fi
  }{}%
}

\pgfplotsset{
  flop clamp curves main/.style={
    flop clamp curves,
    width=7.2cm,
    height=2.5cm,
    every axis x label/.append style={font=\sffamily\scriptsize},
    every axis y label/.append style={font=\sffamily\scriptsize},
    title style={
      font=\sffamily\footnotesize, color=flopInk,
      align=center,
      at={(0.5,1)}, anchor=south,
      yshift=4pt,
      inner ysep=0pt,
    },
  },
}

\pgfplotsset{flop clamp main legend/.style={legend style={at={(axis description cs:0.04,0.85)}, anchor=north west}}}
\newcommand{\flopClampAblationMain}[1]{%
  \begingroup
  \centering
  \begin{tikzpicture}
    \begin{axis}[
      flop clamp curves main,
      title={Humanoid-v4},
      xlabel={Env. Steps (K)},
      ylabel={Episodic Return},
      legend style={
        draw=flopAxis, line width=0.3pt, fill=white, fill opacity=0.9,
        text opacity=1,
        font=\sffamily\scriptsize,
        inner xsep=2pt, inner ysep=2pt,
        row sep=2pt,
        nodes={font=\sffamily\scriptsize, inner ysep=1pt, outer ysep=0pt, inner xsep=1pt},
      },
      flop clamp main legend,
      legend cell align=left,
    ]
      \flopClampAddCurve{#1}{Humanoid-v4}{FQR-noclamp_lr3e-5}{FQR-noclamp}{1}%
      \flopClampAddCurve{#1}{Humanoid-v4}{FQR-clamp_lr3e-5}{FQR-clamp}{1}%
    \end{axis}
  \end{tikzpicture}%
  \endgroup
}

\usetikzlibrary{arrows, arrows.meta}

\pgfplotsset{
  colormap={viridis dark}{
    rgb255=(68,1,84)
    rgb255=(71,39,119)
    rgb255=(61,74,137)
    rgb255=(48,103,141)
    rgb255=(37,131,141)
    rgb255=(30,157,136)
    rgb255=(54,184,119)
    rgb255=(109,206,88)
  },
}

\pgfplotsset{
  flop flowfield/.style={
    width=0.24\linewidth,
    height=2.6cm,
    scale only axis,
    axis line style={draw=flopInk, line width=0.5pt},
    tick style={draw=flopInk, line width=0.4pt},
    tick align=outside,
    tick pos=left,
    major tick length=2pt,
    minor tick length=1pt,
    every tick label/.style={font=\sffamily\scriptsize, color=flopInk},
    title style={
      font=\sffamily\scriptsize, color=flopInk,
      align=center,
      text width=0.32\linewidth,
      at={(0.5,1)}, anchor=south,
      yshift=-2pt,
    },
    label style={font=\sffamily\scriptsize, color=flopInk},
    every axis x label/.style={
      font=\sffamily\tiny, color=flopInk,
      at={(ticklabel cs:0.5)}, anchor=north, yshift=2pt,
    },
    every axis y label/.style={
      font=\sffamily\tiny, color=flopInk,
      at={(ticklabel cs:0.5)}, anchor=south, rotate=90, yshift=-4pt,
    },
    enlarge x limits=false,
    enlarge y limits=false,
    xmin=0, xmax=1,
    ymin=-3, ymax=3,
    xtick={0, 0.5, 1.0},
    xticklabels={0, .5, 1},
    ytick={-3, 0, 3},
    grid=both,
    major grid style={draw=flopAxis!50, line width=0.25pt, dash pattern=on 1pt off 1.5pt},
    minor grid style={draw=flopAxis!30, line width=0.2pt, dash pattern=on 1pt off 1.5pt},
    minor x tick num=1,
    minor y tick num=1,
  },
}

\newcommand{\flopFlowfieldEnvBounds}{%
  \addplot[draw=flopAccentA, line width=0.7pt, dashed, forget plot]
    coordinates {(0, 1) (1, 1)};
  \addplot[draw=flopAccentA, line width=0.7pt, dashed, forget plot]
    coordinates {(0, -1) (1, -1)};
}

\providecommand{\algname}{QF3}

\pgfplotsset{
  flop algo FastTD3/.style ={color=flopAccentB, line width=1.0pt},
  flop algo FPOpp/.style   ={color=flopAccentC, line width=1.0pt},
  flop algo FPOppK48/.style={color=flopAccentC, line width=1.0pt},
  flop band FastTD3/.style ={fill=flopAccentB, fill opacity=0.15, draw=none},
  flop band FPOpp/.style   ={fill=flopAccentC, fill opacity=0.15, draw=none},
  flop band FPOppK48/.style={fill=flopAccentC, fill opacity=0.15, draw=none},
  flop algo FlowRL/.style  ={color=flopAccentH, line width=1.0pt},
  flop band FlowRL/.style  ={fill=flopAccentH, fill opacity=0.15, draw=none},
}

\expandafter\def\csname flopAlgoLabel@FastTD3\endcsname{FastTD3}
\expandafter\def\csname flopAlgoLabel@FPOpp\endcsname{FPO++}
\expandafter\def\csname flopAlgoLabel@FPOppK48\endcsname{FPO++}
\expandafter\def\csname flopAlgoLabel@FlowRL\endcsname{FlowRL}

\tikzset{
  flop loco span/.style={
    draw=flopInk, line width=0.5pt,
    ->, >=stealth,
    shorten >=0pt, shorten <=0pt,
  },
  flop loco span leader/.style={
    draw=flopAxis, line width=0.3pt,
    dash pattern=on 1pt off 1.2pt,
  },
  flop loco span label/.style={
    font=\sffamily\tiny, color=flopInk,
    inner sep=1pt, fill=white, fill opacity=0.85, text opacity=1,
  },
}

\newcommand{\flopAddLocoCurve}[4]{%
  \IfFileExists{#1/locomotion_#2__#3.csv}{%
    \addplot[name path=flopLocoBand#2#3lo, draw=none, forget plot]
      table[x=step, y=lo, col sep=comma]{#1/locomotion_#2__#3.csv};
    \addplot[name path=flopLocoBand#2#3hi, draw=none, forget plot]
      table[x=step, y=hi, col sep=comma]{#1/locomotion_#2__#3.csv};
    \addplot[flop band #3, forget plot]
      fill between[of=flopLocoBand#2#3lo and flopLocoBand#2#3hi];
    \ifnum#4=1
      \addplot+[flop algo #3, mark=none]
        table[x=step, y=mean, col sep=comma]{#1/locomotion_#2__#3.csv};
      \addlegendentry{\csname flopAlgoLabel@#3\endcsname}%
    \else
      \addplot+[flop algo #3, mark=none, forget plot]
        table[x=step, y=mean, col sep=comma]{#1/locomotion_#2__#3.csv};
    \fi
  }{}%
}

\newcommand{\flopAddLocoCurveOrPending}[4]{%
  \IfFileExists{#1/locomotion_#2__#3.csv}{%
    \flopAddLocoCurve{#1}{#2}{#3}{#4}%
  }{%
    \ifnum#4=1
      \addlegendimage{flop algo #3, dashed, mark=none}%
      \addlegendentry{\csname flopAlgoLabel@#3\endcsname{} (pending)}%
    \fi
  }%
}

\pgfplotsset{
  flop loco legend/.style={
    legend pos=south east,
    legend cell align=left,
    legend style={
      draw=flopAxis, line width=0.4pt,
      fill=white, fill opacity=0.85, text opacity=1,
      font=\sffamily\scriptsize,
      inner xsep=2pt, inner ysep=2pt,
      row sep=1.5pt,
      nodes={inner ysep=0.5pt, outer ysep=0pt, font=\sffamily\scriptsize},
    },
  },
}

\expandafter\def\csname flopLocoApplyFastAlgnameLabel\endcsname{%
  \expandafter\def\csname flopAlgoLabel@FLOP\endcsname{\algname{}}%
}

\providecommand{\algname}{QF3}

\pgfplotsset{
  flop splash ref line/.style={color=flopInk!45, line width=0.7pt,
                               dash pattern=on 0.5pt off 1.3pt, line cap=round},
}

\newlength{\flopSplashPairWidth}
\newcommand{\flopSplashPair}[2]{%
  \makebox[\flopSplashPairWidth][l]{%
  \begin{tikzpicture}
    \begin{groupplot}[
      group style={
        group size=2 by 1,
        horizontal sep=6pt,
      },
      flop curves,
      width=0.40\linewidth,
      height=2.8cm,
      every axis x label/.append style={font=\sffamily\scriptsize},
      every axis y label/.append style={font=\sffamily\scriptsize},
      every axis y label/.append style={yshift=2pt},
      every y tick label/.append style={text width=16pt, align=right},
      ylabel style={inner sep=0pt, outer sep=2pt},
      flop algo FastTD3/.style={flop splash ref line},
      flop band FastTD3/.style={draw=none, fill=none},
      #1,
    ]
      #2
    \end{groupplot}
  \end{tikzpicture}%
  }%
}

\pgfplotsset{
  flop splash legend below/.style={
    legend columns=-1,
    legend cell align=left,
    legend style={
      draw=flopAxis, line width=0.4pt, fill=none,
      font=\sffamily\scriptsize,
      column sep=3pt,
      inner xsep=3pt, inner ysep=2pt,
      row sep=1pt,
      nodes={font=\sffamily\scriptsize, inner xsep=1pt, inner ysep=1pt, outer ysep=0pt},
    },
    legend to name={#1},
  },
}
\newcommand{\flopSplashLegendBelow}[1]{%
  \par\vspace{1pt}%
  \makebox[\flopSplashPairWidth][c]{\pgfplotslegendfromname{#1}}%
}

\expandafter\def\csname flopSplashApplyRefLabels\endcsname{%
  \expandafter\def\csname flopAlgoLabel@FastTD3\endcsname{FastTD3 (ref.)}%
}
\newcommand{\flopSplashRefLegend}{%
  \addlegendimage{flop splash ref line, mark=none}%
  \addlegendentry{\csname flopAlgoLabel@FastTD3\endcsname}%
}

\newcommand{\flopSplashVelocityCurves}[1]{%
  \flopSplashPair{}{%
    \nextgroupplot[
      width=0.33\linewidth,
      xmin=0, xmax=5.4e8,
      ymin=0, ymax=0.66,
      ytick={0, 0.2, 0.4, 0.6},
      yticklabel={$\mathsf{\pgfmathprintnumber[fixed, precision=1, 1000 sep={}]{\tick}}$},
      xtick={0, 2.5e8, 5e8},
      xticklabels={0, 250, 500},
      xlabel={Env. Steps (M)},
      ylabel={Lin. Velocity Tracking},
      flop splash legend below=flopSplashLegendVel,
    ]
      \flopAddLocoCurve{#1}{velocity_steps}{FastTD3}{0}%
      \flopAddLocoCurve{#1}{velocity_steps}{FPOppK48}{1}%
      \flopAddLocoCurveOrPending{#1}{velocity_steps}{FlowRL}{1}%
      \flopAddLocoCurve{#1}{velocity_steps}{FLOP}{1}%
      \flopSplashRefLegend
    \nextgroupplot[
      width=0.50\linewidth,
      xmin=0, xmax=10.3,
      ymin=0, ymax=0.66,
      ytick={0, 0.2, 0.4, 0.6},
      xtick={0, 2, 4, 6, 8, 10},
      xticklabels={0, 2, 4, 6, 8, 10},
      xlabel={Wall-clock (hours)},
      yticklabels={},
    ]
      \flopAddLocoCurve{#1}{velocity_hours}{FastTD3}{0}%
      \flopAddLocoCurve{#1}{velocity_hours}{FPOppK48}{0}%
      \flopAddLocoCurve{#1}{velocity_hours}{FLOP}{0}%
  }%
  \flopSplashLegendBelow{flopSplashLegendVel}%
}

\newcommand{\flopSplashMotionCurves}[1]{%
  \flopSplashPair{}{%
    \nextgroupplot[
      width=0.33\linewidth,
      xmode=log,
      xmin=1e7, xmax=4e9,
      ymin=0, ymax=560,
      xtick={1e7, 1e8, 1e9},
      xticklabels={0.01, 0.1, 1},
      log basis x=10,
      xlabel={Env. Steps (B)},
      ylabel={Train Episode Length},
    ]
      \flopAddLocoCurve{#1}{motion_steps}{FastTD3}{0}%
      \flopAddLocoCurve{#1}{motion_steps}{FPOpp}{0}%
      \flopAddLocoCurveOrPending{#1}{motion_steps}{FlowRL}{0}%
      \flopAddLocoCurve{#1}{motion_steps}{FLOP}{0}%
    \nextgroupplot[
      width=0.50\linewidth,
      xmode=log,
      xmin=0.05, xmax=80,
      ymin=0, ymax=560,
      xtick={0.1, 1, 10, 100},
      xticklabels={0.1, 1, 10, 100},
      log basis x=10,
      xlabel={Wall-clock (hours)},
      yticklabels={},
      flop splash legend below=flopSplashLegendMot,
    ]
      \flopAddLocoCurve{#1}{motion_hours}{FastTD3}{0}%
      \flopAddLocoCurve{#1}{motion_hours}{FPOpp}{1}%
      \flopAddLocoCurveOrPending{#1}{motion_hours}{FlowRL}{1}%
      \flopAddLocoCurve{#1}{motion_hours}{FLOP}{1}%
      \flopSplashRefLegend
      \draw[flop loco span leader] (axis cs:0.88,480) -- (axis cs:0.88,545);
      \draw[flop loco span leader] (axis cs:8.93,480) -- (axis cs:8.93,545);
      \draw[flop loco span] (axis cs:8.93,545) -- (axis cs:0.88,545);
      \node[flop loco span label, right] at (axis cs:8.93,545) {$10\times$};%
  }%
  \flopSplashLegendBelow{flopSplashLegendMot}%
}

\newcommand{\flopLocomotionResultsOneRow}[1]{%
  \begingroup
  \flopLocoApplyFastAlgnameLabel
  \flopSplashApplyRefLabels
  \renewcommand{\flopSplashLegendBelow}[1]{}%
  \begin{subfigure}[b]{0.49\linewidth}\raggedright
    \setlength{\flopSplashPairWidth}{\linewidth}%
    \flopSplashVelocityCurves{#1}%
    \caption{G1 linear-velocity tracking.}
    \label{fig:splash-vel-curves}
  \end{subfigure}\hfill
  \begin{subfigure}[b]{0.49\linewidth}\raggedright
    \setlength{\flopSplashPairWidth}{\linewidth}%
    \flopSplashMotionCurves{#1}%
    \caption{LAFAN dance1\_subject2 single-motion tracking.}
    \label{fig:splash-mot-curves}
  \end{subfigure}\par\vspace{2pt}
  \centering\pgfplotslegendfromname{flopSplashLegendVel}%
  \endgroup
}

\pgfplotsset{
  flop flowfield pair/.style={
    flop flowfield,
    width=2.9cm,
    height=2.3cm,
    every axis x label/.append style={font=\sffamily\scriptsize},
    every axis y label/.append style={font=\sffamily\scriptsize},
    title style={
      font=\sffamily\footnotesize, color=flopInk,
      text width=,
      at={(0.5,1)}, anchor=south,
      yshift=1pt,
      inner ysep=0pt,
    },
  },
}

\newcommand{\flopFlowfieldStreamsPair}[2]{%
  \addplot[
    mesh, line width=0.95pt,
    point meta=explicit,
    unbounded coords=jump,
    forget plot,
  ]
    table[x=tau, y=action, meta=speed, col sep=comma]{#1/flow_#2.csv};
}
\newcommand{\flopFlowfieldArrowsPair}[2]{%
  \addplot[
    point meta=explicit,
    quiver={u=\thisrow{dx}, v=\thisrow{dy},
            every arrow/.append style={
              -{Triangle[length=2.4pt, width=2.0pt]},
              draw=mapped color, fill=mapped color,
              line width=0.05pt,
            }},
    forget plot,
  ]
    table[x=x, y=y, meta=speed, col sep=comma]{#1/arrows_#2.csv};
}

\definecolor{flopBendMark}{HTML}{DC2626}

\newcommand{\flopFlowfieldClampRow}[3]{%
  \begingroup
  \centering
  \begin{tikzpicture}
    \begin{groupplot}[
      group style={
        group size=2 by 1,
        horizontal sep=10pt,
        ylabels at=edge left,
        yticklabels at=edge left,
      },
      flop flowfield pair,
      width=#2, height=#3,
      colormap name=viridis dark,
      point meta min=0, point meta max=18,
      yticklabel={$\mathsf{\pgfmathprintnumber{\tick}}$},
    ]
      \nextgroupplot[title={Velocity clamp}, xlabel={$\tau$}]
      \flopFlowfieldStreamsPair{#1}{velocity}%
      \flopFlowfieldArrowsPair{#1}{velocity}%
      \flopFlowfieldEnvBounds
      \nextgroupplot[title={Action clamp}, xlabel={$\tau$}, clip mode=individual]
      \flopFlowfieldStreamsPair{#1}{action_best}%
      \flopFlowfieldArrowsPair{#1}{action_best}%
      \flopFlowfieldEnvBounds
      \draw[flopBendMark, line width=0.9pt, dash pattern=on 2pt off 1.5pt]
        (axis cs:0.94,-1.9) ellipse [x radius=0.10, y radius=1.2];
    \end{groupplot}
    \begin{axis}[
      hide axis,
      scale only axis, height=#3, width=0pt,
      at={($(group c2r1.south east)+(4pt,0)$)}, anchor=south west,
      colormap name=viridis dark,
      point meta min=0, point meta max=18,
      colorbar,
      colorbar style={
        width=3pt,
        ytick={0, 6, 12, 18},
        yticklabel={$\mathsf{\pgfmathprintnumber{\tick}}$},
        title={$|v|$},
        title style={font=\scriptsize, color=flopInk, at={(0.5,1)}, anchor=south, yshift=-3.7pt, inner ysep=0pt},
        every tick label/.style={font=\sffamily\tiny, color=flopInk},
        axis line style={draw=flopInk, line width=0.4pt},
      },
    ]
      \addplot[draw=none] coordinates {(0,0) (1,1)};
    \end{axis}
  \end{tikzpicture}%
  \endgroup
}

\pgfplotsset{
  manipulation algo QF/.style       ={color=flopAccentA, line width=1.5pt, solid},
  manipulation algo NoClamp/.style  ={color=flopAccentA!45!white, line width=1.0pt, solid},
  manipulation algo Residual/.style ={color=flopAccentB, line width=1.0pt, solid},
  manipulation algo Base/.style     ={color=flopInk!65, line width=0.7pt, dash pattern=on 2.5pt off 1.5pt},
  manipulation band QF/.style       ={fill=flopAccentA, fill opacity=0.18, draw=none},
  manipulation band NoClamp/.style  ={fill=flopAccentA, fill opacity=0.07, draw=none},
  manipulation band Residual/.style ={fill=flopAccentB, fill opacity=0.15, draw=none},
  manipulation external legend/.style args={#1}{
    legend to name=#1,
    legend columns=4,
    legend style={
      draw=flopAxis, line width=0.4pt, fill=none,
      font=\sffamily\scriptsize,
      legend cell align=left,
      column sep=3pt,
      inner xsep=3pt, inner ysep=2pt,
      row sep=1pt,
      nodes={font=\sffamily\scriptsize, inner xsep=1pt, inner ysep=1pt, outer ysep=0pt},
    },
  },
}

\newcommand{\flopAddManipCurve}[4]{%
  \IfFileExists{#1/manipulation_#2_#3__#4.csv}{%
    \addplot[name path=flopManipBand#2#3#4lo, draw=none, forget plot]
      table[x=step, y=lo, col sep=comma]{#1/manipulation_#2_#3__#4.csv};
    \addplot[name path=flopManipBand#2#3#4hi, draw=none, forget plot]
      table[x=step, y=hi, col sep=comma]{#1/manipulation_#2_#3__#4.csv};
    \addplot[manipulation band #4, forget plot]
      fill between[of=flopManipBand#2#3#4lo and flopManipBand#2#3#4hi];
    \addplot+[manipulation algo #4, mark=none, forget plot]
      table[x=step, y=mean, col sep=comma]{#1/manipulation_#2_#3__#4.csv};
  }{}%
}

\newcommand{\flopAddManipLine}[4]{%
  \IfFileExists{#1/manipulation_#2_#3__#4.csv}{%
    \addplot+[manipulation algo #4, mark=none, forget plot]
      table[x=step, y=mean, col sep=comma]{#1/manipulation_#2_#3__#4.csv};
  }{}%
}

\newcommand{\flopAddManipBase}[2]{%
  \ifcsname flopManipBase@#1@#2\endcsname
    \edef\flopManipBaseVal{\csname flopManipBase@#1@#2\endcsname}%
    \draw[/pgfplots/manipulation algo Base]
      ({rel axis cs:0,0}|-{axis cs:0,\flopManipBaseVal}) --
      ({rel axis cs:1,0}|-{axis cs:0,\flopManipBaseVal});
  \fi
}

\newcommand{\flopManipPendingNote}[2]{%
  \IfFileExists{#1/manipulation_#2_success__QF.csv}{}{%
    \node[font=\sffamily\tiny, text=flopInk!55, fill=white, inner sep=2pt]
      at (axis description cs:0.5,0.5) {Results pending};
  }%
}

\newlength{\flopManipH}          
\newlength{\flopManipStillW}     
\newlength{\flopManipAxisW}      
\newlength{\flopManipGapStill}   
\newlength{\flopManipGapAxes}    
\newlength{\flopManipLabelW}     
\pgfplotsset{
  manipulation grid axis/.style={
    flop curves,
    scale only axis,
    width=\flopManipAxisW,
    xmin=0, xmax=1,
    xtick={0,0.5,1},
    xticklabels={,,},
    xlabel={},
    every axis x label/.append style={font=\sffamily\scriptsize},
    every y tick label/.append style={text width=13pt},
  },
}

\newcommand{\flopManipStills}[2]{%
  \node[anchor=north east, inner sep=0pt] (flopStill)
    at (-\flopManipGapStill, 0) {#2};
  \node[rotate=90, anchor=south, inner sep=0pt, font=\sffamily\scriptsize, text=flopInk]
    at ([xshift=-3pt]flopStill.west) {#1};
}
\newcommand{\flopManipStill}[1]{\includegraphics[width=\flopManipStillW, height=\flopManipH]{#1}}
\newcommand{\flopManipHeaders}{%
  \begin{scope}[every node/.style={anchor=south, inner sep=0pt, yshift=3pt,
                                   font=\sffamily\scriptsize, text=flopInk}]
    \node at (0.5\flopManipAxisW, 0) {Success Rate (\%) $\uparrow$};
    \node at (1.5\flopManipAxisW+\flopManipGapAxes, 0) {Mean Succ.\ Ep.\ Length $\downarrow$};
    \node at (2.5\flopManipAxisW+2\flopManipGapAxes, 0) {Successful Within Time (\%) $\uparrow$};
  \end{scope}
}
\tikzset{
  flop manip row/.style={
    trim left={-(\flopManipGapStill+\flopManipStillW+3pt+\flopManipLabelW)},
    trim right={3\flopManipAxisW+2\flopManipGapAxes+4pt},
  },
}

\newcommand{\flopManipSuccessAxis}[4]{%
  \begin{axis}[
    manipulation grid axis,
    at={(0,0)}, anchor=north west,
    height=\flopManipH,
    ymin=0, ymax=1, ytick={0,0.5,1}, yticklabels={0,50,100},
    #3, #4,
  ]
    \flopAddManipBase{#2}{success}%
    \flopAddManipCurve{#1}{#2}{success}{NoClamp}%
    \flopAddManipCurve{#1}{#2}{success}{Residual}%
    \flopAddManipCurve{#1}{#2}{success}{QF}%
    \flopManipPendingNote{#1}{#2}%
  \end{axis}
}

\newcommand{\flopManipSuccessAxisBroken}[5]{%
  \pgfmathsetlengthmacro{\flopBreakGap}{4pt}%
  \pgfmathsetlengthmacro{\flopBreakUpH}{(\flopManipH-\flopBreakGap)*0.25/0.37}%
  \pgfmathsetlengthmacro{\flopBreakLoH}{(\flopManipH-\flopBreakGap)*0.12/0.37}%
  \begin{axis}[
    manipulation grid axis,
    name=flopSuccUp,
    at={(0,0)}, anchor=north west,
    height=\flopBreakUpH,
    #3, #4,
    xticklabels={}, xlabel={},
    x tick style={draw=none}, minor x tick style={draw=none},
    axis line style={draw=none},
  ]
    \flopAddManipBase{#2}{success}%
    \flopAddManipLine{#1}{#2}{success}{NoClamp}%
    \flopAddManipCurve{#1}{#2}{success}{Residual}%
    \flopAddManipCurve{#1}{#2}{success}{QF}%
  \end{axis}
  \path (flopSuccUp.south west) ++(0,-\flopBreakGap) coordinate (flopSuccLoAnchor);
  \begin{axis}[
    manipulation grid axis,
    name=flopSuccLo,
    at={(flopSuccLoAnchor)}, anchor=north west,
    height=\flopBreakLoH,
    axis line style={draw=none},
    #3, #5,
  ]
    \flopAddManipLine{#1}{#2}{success}{NoClamp}%
    \flopAddManipCurve{#1}{#2}{success}{Residual}%
    \flopAddManipCurve{#1}{#2}{success}{QF}%
  \end{axis}
  \draw[flopInk, line width=0.5pt]
    (flopSuccUp.south west) -- (flopSuccUp.north west) -- (flopSuccUp.north east) -- (flopSuccUp.south east)
    (flopSuccLo.north west) -- (flopSuccLo.south west) -- (flopSuccLo.south east) -- (flopSuccLo.north east);
  \draw[flopInk!35, line width=0.4pt]
    (flopSuccUp.south west) -- (flopSuccUp.south east)
    (flopSuccLo.north west) -- (flopSuccLo.north east);
  \draw[flopAxis, line width=0.5pt]
    ([xshift=-1.5pt, yshift=-1.2pt]flopSuccUp.south west) --
    ([xshift=1.5pt, yshift=1.2pt]flopSuccUp.south west)
    ([xshift=-1.5pt, yshift=-1.2pt]flopSuccLo.north west) --
    ([xshift=1.5pt, yshift=1.2pt]flopSuccLo.north west);
}

\newcommand{\flopManipEplenAxis}[4]{%
  \begin{axis}[
    manipulation grid axis,
    at={(\flopManipAxisW+\flopManipGapAxes, 0)}, anchor=north west,
    height=\flopManipH,
    ymin=0, ymax=1, ytick={0,0.5,1}, yticklabels={,,},
    every y tick label/.append style={text width=20pt},
    #3, #4,
  ]
    \flopAddManipBase{#2}{eplen}%
    \flopAddManipCurve{#1}{#2}{eplen}{NoClamp}%
    \flopAddManipCurve{#1}{#2}{eplen}{Residual}%
    \flopAddManipCurve{#1}{#2}{eplen}{QF}%
    \flopManipPendingNote{#1}{#2}%
  \end{axis}
}

\newcommand{\flopAddManipTime}[3]{%
  \IfFileExists{#1/manipulation_#2_bytime__#3.csv}{%
    \addplot+[manipulation algo #3, line width=1.1pt, const plot, mark=none, forget plot,
              name path=flopManipTime#3]
      table[x=t, y=rate, col sep=comma]{#1/manipulation_#2_bytime__#3.csv};
  }{}%
}
\newcommand{\flopManipTimeAxis}[3]{%
  \begin{axis}[
    manipulation grid axis,
    at={(2\flopManipAxisW+2\flopManipGapAxes, 0)}, anchor=north west,
    height=\flopManipH,
    ymin=0, ymax=1, ytick={0,0.5,1}, yticklabels={0,50,100},
    enlarge x limits=false,
    #3,
  ]
    \flopAddManipTime{#1}{#2}{Base}%
    \flopAddManipTime{#1}{#2}{QF}%
    \addplot[fill=flopAccentA, fill opacity=0.16, draw=none, forget plot]
      fill between[of=flopManipTimeQF and flopManipTimeBase];
  \end{axis}
}

\newcommand{\flopManipLegend}{%
  \begin{tikzpicture}
    \begin{axis}[hide axis, scale only axis, width=1cm, height=1cm, xmin=0, xmax=1, ymin=0, ymax=1,
                 manipulation external legend={manipulationLegend}]
      \addlegendimage{manipulation algo QF}\addlegendentry{\algname{}}
      \addlegendimage{manipulation algo NoClamp}\addlegendentry{Direct Q (unclamped)}
      \addlegendimage{manipulation algo Residual}\addlegendentry{Residual RL}
      \addlegendimage{manipulation algo Base}\addlegendentry{Base}
    \end{axis}
  \end{tikzpicture}%
  \pgfplotslegendfromname{manipulationLegend}%
}

\newcommand{\flopManipulationFinetuning}[1]{%
  \begingroup
  \IfFileExists{#1/manipulation_base.tex}{\input{#1/manipulation_base.tex}}{}%
  \centering
  \phantomsubcaption\label{fig:manipulation-bottles}%
  \begin{tikzpicture}[flop manip row]
    \flopManipStills{(\thesubfigure) Bottles}
      {\flopManipStill{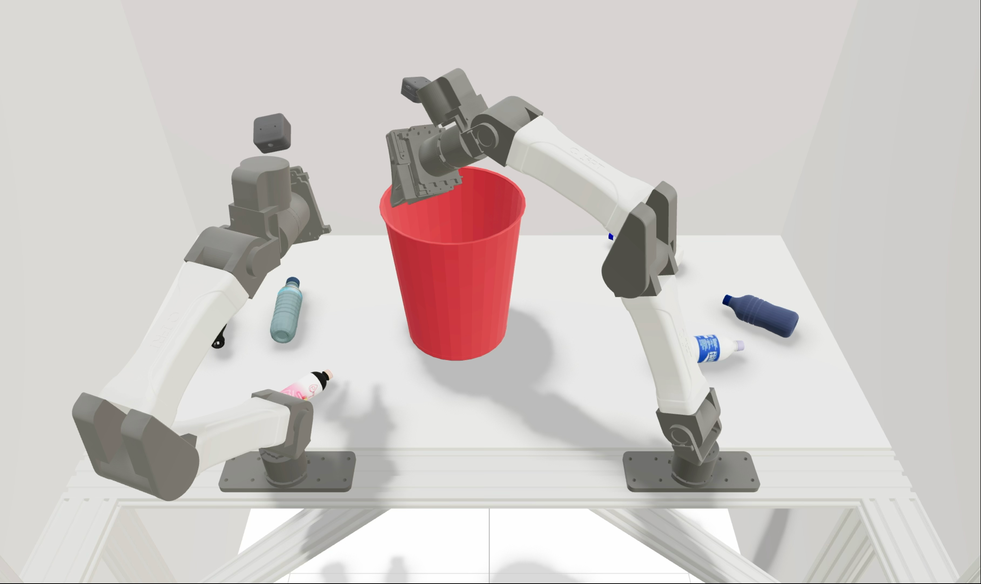}}
    \flopManipHeaders
    \flopManipSuccessAxisBroken{#1}{bottles}{%
      xmin=0, xmax=4.2e5, xtick={0, 2e5, 4e5}, xticklabels={0, 200, 400},
    }{%
      ymin=0.75, ymax=1.0, enlarge y limits=0,
      ytick={0.8, 0.9, 1.0}, yticklabels={80, 90, 100},
    }{%
      ymin=-0.02, ymax=0.10, enlarge y limits=0,
      ytick={0, 0.1}, yticklabels={0, 10},
    }
    \flopManipEplenAxis{#1}{bottles}{%
      xmin=0, xmax=4.2e5, xtick={0, 2e5, 4e5}, xticklabels={0, 200, 400},
    }{%
      ymin=550, ymax=850, ytick={600, 700, 800}, yticklabels={600, 700, 800},
    }
    \flopManipTimeAxis{#1}{bottles}{%
      xmin=0, xmax=34, xtick={0, 15, 30}, xticklabels={0, 15, 30},
    }
  \end{tikzpicture}\par\vspace{5pt}
  \phantomsubcaption\label{fig:manipulation-dishrack}%
  \begin{tikzpicture}[flop manip row]
    \flopManipStills{(\thesubfigure) Dishrack}
      {\flopManipStill{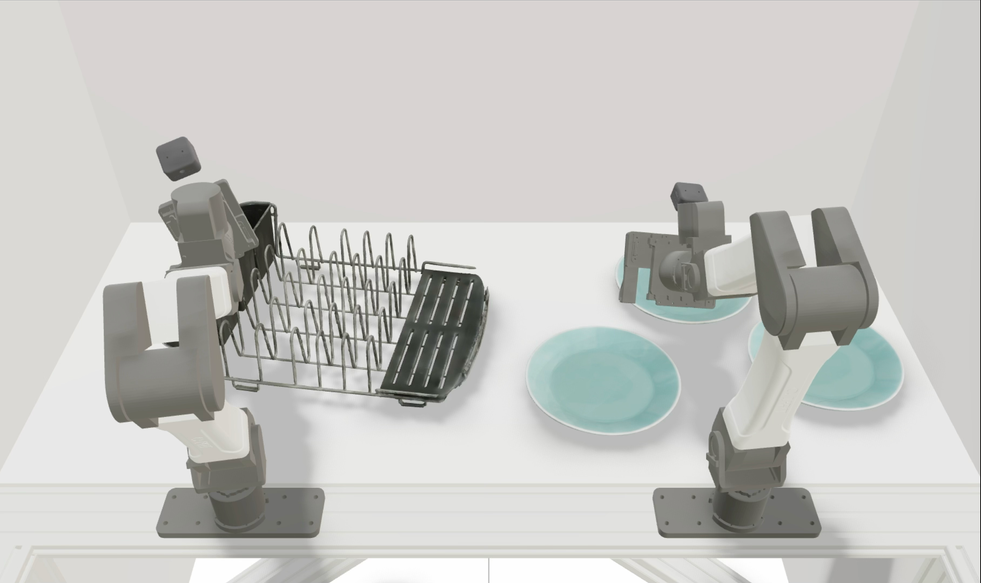}}
    \flopManipSuccessAxis{#1}{dishrack}{%
      xmin=0, xmax=5.6e5, xtick={0, 2.5e5, 5e5}, xticklabels={0, 250, 500},
      xlabel={Env. Steps (k)},
    }{}
    \flopManipEplenAxis{#1}{dishrack}{%
      xmin=0, xmax=5.6e5, xtick={0, 2.5e5, 5e5}, xticklabels={0, 250, 500},
      xlabel={Env. Steps (k)},
    }{%
      ymin=800, ymax=2400, ytick={1000, 1500, 2000}, yticklabels={1000, 1500, 2000},
    }
    \flopManipTimeAxis{#1}{dishrack}{%
      xmin=0, xmax=112, xtick={0, 50, 100}, xticklabels={0, 50, 100},
      xlabel={Time (s)},
    }
  \end{tikzpicture}\par\vspace{4pt}
  \flopManipLegend\par
  \endgroup
}

\pgfplotsset{
  robomimic algo QF/.style      ={color=flopAccentA, line width=1.1pt, dash pattern=on 3pt off 1.5pt},
  robomimic algo QFPlus/.style  ={color=flopAccentA, line width=1.6pt, solid},
  robomimic algo OGPO/.style    ={color=flopAccentB, line width=1.0pt, solid},
  robomimic algo OGPOPlus/.style={color=flopAccentB!50!white, line width=1.0pt, solid},
  robomimic algo OGPOCA/.style  ={color=flopAccentB, line width=1.0pt, solid},
  robomimic algo DSRL/.style    ={color=flopAccentC, line width=1.0pt, solid},
  robomimic algo EXPO/.style    ={color=flopAccentD, line width=1.0pt, solid},
  robomimic band QF/.style      ={fill=flopAccentA, fill opacity=0.12, draw=none},
  robomimic band QFPlus/.style  ={fill=flopAccentA, fill opacity=0.18, draw=none},
  robomimic band OGPO/.style    ={fill=flopAccentB, fill opacity=0.15, draw=none},
  robomimic band OGPOPlus/.style={fill=flopAccentB, fill opacity=0.07, draw=none},
  robomimic band OGPOCA/.style  ={fill=flopAccentB, fill opacity=0.12, draw=none},
  robomimic band DSRL/.style    ={fill=flopAccentC, fill opacity=0.15, draw=none},
  robomimic band EXPO/.style    ={fill=flopAccentD, fill opacity=0.15, draw=none},
}
\newif\ifflopRobomimicHaveData
\newcommand{\flopAddRobomimicCurve}[3]{%
  \IfFileExists{#1/robomimic_#2_success__#3.csv}{%
    \global\flopRobomimicHaveDatatrue
    \addplot[name path=flopRobomimicBand#2#3lo, draw=none, forget plot]
      table[x=step, y=lo, col sep=comma]{#1/robomimic_#2_success__#3.csv};
    \addplot[name path=flopRobomimicBand#2#3hi, draw=none, forget plot]
      table[x=step, y=hi, col sep=comma]{#1/robomimic_#2_success__#3.csv};
    \addplot[robomimic band #3, forget plot]
      fill between[of=flopRobomimicBand#2#3lo and flopRobomimicBand#2#3hi];
    \addplot+[robomimic algo #3, mark=none, forget plot]
      table[x=step, y=mean, col sep=comma]{#1/robomimic_#2_success__#3.csv};
  }{}%
}
\newcommand{\flopRobomimicPanel}[4]{%
  \begin{subfigure}[t]{0.32\textwidth}
    \centering
    \begin{tikzpicture}
      \begin{axis}[
        flop curves,
        scale only axis,
        width=\dimexpr\linewidth-32pt\relax,
        height=2.4cm,
        ymin=0, ymax=1,
        ytick={0,0.5,1}, yticklabels={0,50,100},
        xlabel={Env. Steps (M)},
        ylabel={Success Rate (\%)},
        every axis x label/.append style={font=\sffamily\scriptsize},
        every axis y label/.append style={font=\sffamily\scriptsize},
        #4,
      ]
        \global\flopRobomimicHaveDatafalse
        \flopAddRobomimicCurve{#1}{#2}{EXPO}%
        \flopAddRobomimicCurve{#1}{#2}{DSRL}%
        \flopAddRobomimicCurve{#1}{#2}{OGPOCA}%
        \flopAddRobomimicCurve{#1}{#2}{OGPOPlus}%
        \flopAddRobomimicCurve{#1}{#2}{OGPO}%
        \flopAddRobomimicCurve{#1}{#2}{QFPlus}%
        \flopAddRobomimicCurve{#1}{#2}{QF}%
        \ifflopRobomimicHaveData\else
          \node[font=\sffamily\tiny, text=flopInk!55, fill=white, inner sep=2pt]
            at (axis description cs:0.5,0.5) {Results pending};
        \fi
      \end{axis}
    \end{tikzpicture}
    \caption{\textup{\texttt{#3}}}
  \end{subfigure}%
}
\newcommand{\flopRobomimicLegend}{%
  \begin{tikzpicture}
    \begin{axis}[hide axis, scale only axis, width=1cm, height=1cm, xmin=0, xmax=1, ymin=0, ymax=1,
                 manipulation external legend={robomimicLegend}, legend columns=5]
      \addlegendimage{robomimic algo QF}\addlegendentry{\algname{}}
      \addlegendimage{robomimic algo QFPlus}\addlegendentry{\algname{}+}
      \addlegendimage{robomimic algo OGPOCA}\addlegendentry{OGPO+CA}
      \addlegendimage{robomimic algo DSRL}\addlegendentry{DSRL}
      \addlegendimage{robomimic algo EXPO}\addlegendentry{EXPO}
    \end{axis}
  \end{tikzpicture}%
  \pgfplotslegendfromname{robomimicLegend}%
}
\newcommand{\flopRobomimicOdeLegend}{%
  \begin{tikzpicture}
    \begin{axis}[hide axis, scale only axis, width=1cm, height=1cm, xmin=0, xmax=1, ymin=0, ymax=1,
                 manipulation external legend={robomimicOdeLegend}, legend columns=4]
      \addlegendimage{robomimic algo QF}\addlegendentry{\algname{}}
      \addlegendimage{robomimic algo QFPlus}\addlegendentry{\algname{}+}
      \addlegendimage{robomimic algo OGPOCA}\addlegendentry{OGPO+CA}
      \addlegendimage{robomimic algo EXPO}\addlegendentry{EXPO}
    \end{axis}
  \end{tikzpicture}%
  \pgfplotslegendfromname{robomimicOdeLegend}%
}
\newcommand{\flopRobomimicPanels}[1]{%
  \centering
  \flopRobomimicPanel{#1}{square}{square-mh-low\_dim-clip}{%
    xmin=0, xmax=1.5e6, xtick={0,0.5e6,1e6,1.5e6}, xticklabels={0,0.5,1,1.5}}\hfill
  \flopRobomimicPanel{#1}{toolhang}{tool\_hang-ph-low\_dim}{%
    xmin=0, xmax=2e6, xtick={0,1e6,2e6}, xticklabels={0,1,2}}\hfill
  \flopRobomimicPanel{#1}{transport}{transport-mh-low\_dim}{%
    xmin=0, xmax=2e6, xtick={0,1e6,2e6}, xticklabels={0,1,2}}\par\vspace{4pt}
}
\newcommand{\flopRobomimic}[1]{\flopRobomimicPanels{#1}\flopRobomimicLegend\par}
\newcommand{\flopRobomimicOde}[1]{\flopRobomimicPanels{#1}\flopRobomimicOdeLegend\par}

\definecolor{qfLink}{HTML}{006666}
\usepackage[colorlinks=true,linkcolor=qfLink,citecolor=qfLink,urlcolor=qfLink]{hyperref}
\hypersetup{pdftitle={Fast Off-Policy RL with Filtered Q-Gradients for Flow Policies}}

\newcommand{\qfTitle}{\algname{}: Fast Flow RL with Filtered Q-Gradients}
\newcommand{\qfAuthors}{%
  Chung~Min~Kim$^{1,2}$,
  Brent~Yi$^{1,2}$,
  David~McAllister$^{1,2}$,
  Hongsuk~Choi$^{1,2}$,
  Himanshu~Gaurav~Singh$^{1}$,
  Jinkun~Cao$^{2}$,
  Ken~Goldberg$^{1}$,
  Pieter~Abbeel$^{1,2}$,
  Carmelo~Sferrazza$^{2\dagger}$,
  Angjoo~Kanazawa$^{1,2\dagger}$}
\newcommand{\qfAffiliations}{%
  $^{1}$UC Berkeley \quad $^{2}$Amazon FAR \quad $^{\dagger}$FAR Team Co-lead \quad
  Work done as an intern at Amazon FAR}

\makeatletter
\newcommand{\qfMakeTitle}{%
  \thispagestyle{plain}%
  \begingroup\raggedright
    \vspace*{-0.1in}%
    {\fontsize{24}{28}\selectfont\bfseries \qfTitle\par}
    \vspace{12pt}%
    {\normalsize \qfAuthors\par}
    \vspace{12pt}%
  \endgroup
  \ifdefempty{\qfAffiliations}{}{%
    \begingroup
      \renewcommand{\thefootnote}{}%
      \renewcommand{\@makefntext}[1]{\noindent##1}%
      \footnotetext{\qfAffiliations}%
    \endgroup}}
\makeatother

\begin{document}
\qfMakeTitle

\noindent\begin{minipage}{\textwidth}
    \centering
    \captionsetup{type=figure}
    \includegraphics[width=\textwidth]{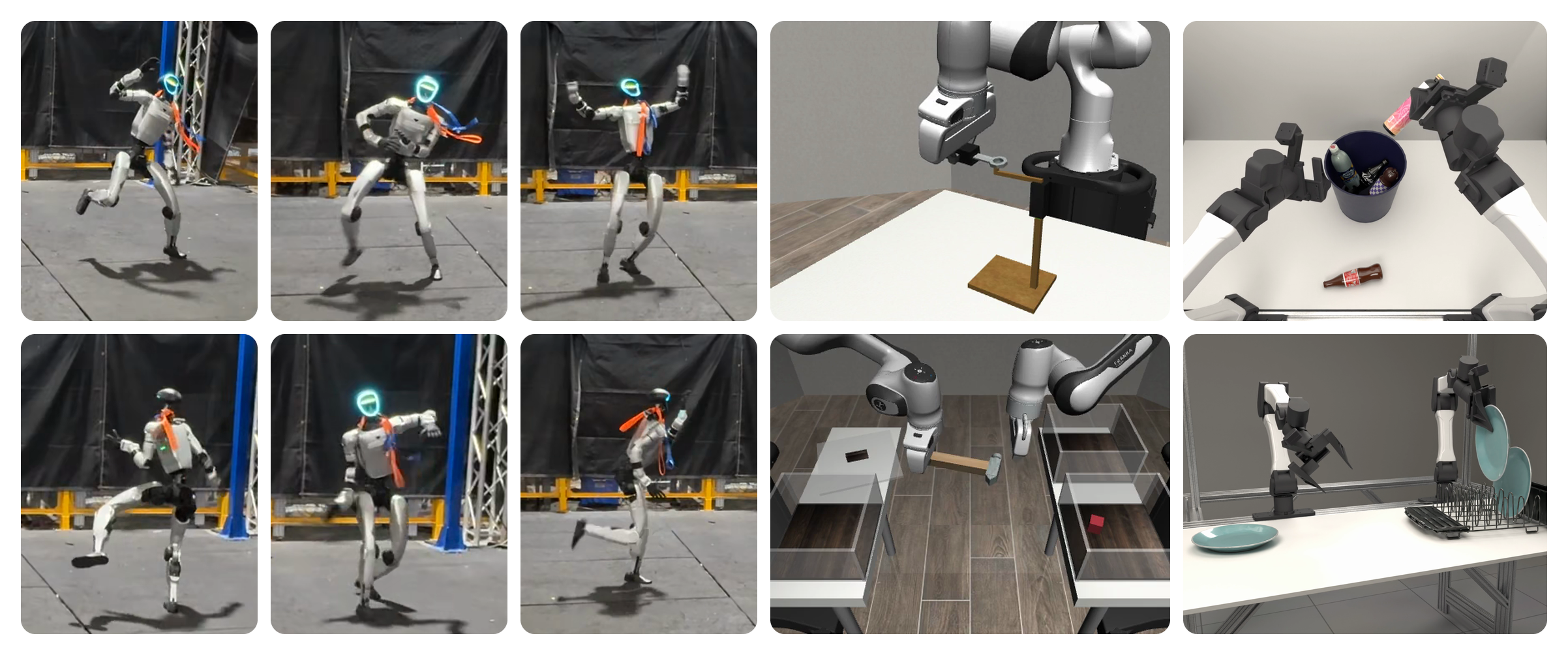}
    \caption{
    \textbf{\algname{} is an off-policy RL algorithm for training flow policies online, either from scratch or starting from pretrained policies.}
    \textit{Left:} A Unitree G1 humanoid tracking a $\sim$3 minute dynamic dance motion
    with a whole-body motion-tracking policy trained from scratch in simulation with
    \algname{} and deployed zero-shot on hardware. Training is an order of magnitude
    faster in wall-clock time than recent on-policy flow baselines.
    \textit{Right:} \algname{} improves manipulation performance by fine-tuning
    flow policies online in simulation (Robomimic tool hang and transport;
    ABC-sim gathering bottles, loading a dishrack).}
    \label{fig:humanoid-sim2real-curves}
\end{minipage}
\par\vspace{12pt}

\noindent\textbf{Abstract.}~\ignorespaces
Flow policies have become a standard policy class for learning robot behaviors from demonstrations, but reinforcement learning is still critical for improving pre-trained flow policies or learning them from scratch through interaction.
We introduce \algname{}
(Fast Flow RL with Filtered Q-Gradients),
an
online off-policy RL algorithm that trains a flow policy with flow matching plus the critic's action gradient, backpropagated through a one-step prediction of the flow's output. To keep updates where the critic and this prediction are reliable, \algname{} applies the critic gradient only to action dimensions that stay near the replay action.
To our knowledge, \algname{} is the first off-policy flow RL method
to train humanoid locomotion policies from scratch and transfer them
zero-shot to hardware. Paired with a high-throughput off-policy training
recipe, it trains humanoid locomotion and motion-tracking policies
with a 10$\times$ wall-clock speedup over FPO++, a recent on-policy
flow RL method.
We further apply \algname{} to fine-tune pretrained
flow-based manipulation policies
on both ABC-Sim and Robomimic tasks.
These results suggest that \algname{} can both learn
robot policies from scratch and refine those acquired from
demonstrations.
Website: \url{https://qf3-rl.github.io/}.

\par\vspace{6pt}

\section{Introduction}

Flow and diffusion policies are widely used in robot learning.
They are the dominant approach for learning manipulation skills from demonstrations, including in large pretrained robot policies~\citep{black2024pi0,intelligence2025pi05,tri2025lbm,nvidia2025groot,jang2025dreamgen}, and are increasingly adopted for locomotion and whole-body control~\citep{liao2025beyondmimic,wang2026omnixtreme}.

Training flow policies by imitation alone is often insufficient: for some tasks, expert trajectories may be unavailable and the policy must learn from scratch through rewards; for others, a pretrained policy provides an initial behavior that can be improved through interaction, e.g., to complete a task more reliably or quickly.
Settings like these have motivated a growing body of reinforcement-learning methods for flow policies, including both policy-gradient and value-based approaches~\citep{mcallister2025fpo,yi2026fpopp,yang2023dipo,lv2025flowbased,shi2026flowdpg}, both for learning from scratch~\citep{mcallister2025fpo,yi2026fpopp,yang2023dipo,psenka2023qsm,lv2025flowbased} and for fine-tuning pretrained policies~\citep{dong2025expo,patil2026ogpo,shi2026flowdpg}.

Off-policy learning is particularly attractive for robotics: not only can it reduce the need for additional interactions by reusing collected experience, but it can also support fast wall-clock training when combined with parallel simulation and large-batch updates~\citep{seo2025fasttd3}.
TD3-style actors follow the critic's action gradient $\nabla_a Q$, which gives a per-dimension direction of improvement for each action. For a flow policy, however, this means backpropagating through every step of the sampler, which is costly and can be unstable~\citep{wang2023diffusionql,wang2024dacer,zhang2025sacflow,park2025fql}.
Existing methods for off-policy flow RL therefore either pay this cost and differentiate through the sampler~\citep{wang2023diffusionql,wang2024dacer,zhang2025sacflow,lv2025flowbased}, freeze the flow and train a residual or noise policy on top~\citep{wang2026omnixtreme,dong2025expo,wagenmaker2025steering}, or use the critic only through its values or as regression targets~\citep{patil2026ogpo,lei2025rl100,psenka2023qsm,yang2023dipo}.

We introduce \algname{}, Fast, Filtered Q-learning for Flow policies, a general online off-policy RL algorithm for flow policies. In contrast to approaches that trains a residual policy~\citep{wang2026omnixtreme,dong2025expo,xiao2025self} or modify the sampling inputs of
a fixed pretrained flow~\citep{wagenmaker2025steering,xu2026expertgen}, QF3 directly trains a flow policy, even from scratch, in a stable and sample efficient manner.
It is designed to drop into existing TD3-style off-policy training recipes, enabling
efficient training in large-scale parallel simulation.
Its actor objective is simple: it combines standard flow matching with $Q$-maximization through a one-step prediction of the final action, and can be evaluated directly on replay
samples, without integrating or differentiating through the flow ODE.

Concretely, QF3 backpropagates a $Q$ maximization objective through the learned velocity field, while stabilizing updates by clipping $Q$-induced velocity corrections around the replay-conditioned flow target.
This forms a replay-local trust region: velocity coordinates whose $Q$-induced correction would move too far from the replay-conditioned flow target receive no critic gradient. $Q$-maximization therefore stays near replay actions, where the critic has been trained and the one-step prediction is accurate.

We evaluate \algname{}'s effectiveness both for training flow policies from scratch and for finetuning pretrained policies, on humanoid locomotion and manipulation respectively. Our contributions are as follows:

\textbf{1. Algorithm and analysis.} We introduce \algname{}, a simple off-policy actor update for flow policies, and provide an analysis of how its velocity clip filters critic-driven updates for stable off-policy learning (Sec.~\ref{sec:algorithm}, \ref{sec:analysis}). On MuJoCo, results suggest that \algname{}, with just a single clipped-gradient update and no sampler backpropagation, performs competitively with recent off-policy flow RL methods (Sec.~\ref{sec:exp:mujoco}).

\textbf{2. Scaling to high-dimensional, dynamic control from scratch.} We evaluate \algname{} on velocity- and motion-tracking for the 29-DoF Unitree G1, training from scratch in simulation and deploying zero-shot on hardware; to our knowledge, this is the first off-policy flow RL method shown to do so. Our results show wall-clock training times close to FastTD3 and an order of magnitude below FPO++, an on-policy flow RL method (Sec.~\ref{sec:exp:locomotion}).

\textbf{3. Manipulation fine-tuning.} We apply \algname{} to fine-tune pretrained flow policies in ABC-Sim~\citep{abc2026} and Robomimic~\citep{mandlekar2021robomimic}. In ABC-Sim, we fine-tune the flow action head of the ABC-VLA policy with LoRA on two bimanual tasks; on gathering bottles into a bin, the fine-tuned policy completes episodes 22\% faster than the base policy, and its success rate rises from 91\% to 95\% (Sec.~\ref{sec:exp:manipulation}). On three Robomimic tasks, where we fine-tune the full policy, results show \algname{} performing similarly or better compared to OGPO~\citep{patil2026ogpo}(Sec.~\ref{sec:exp:robomimic}).

\section{Related Work}
\label{sec:related_work}

\subsection{Diffusion and Flow Models for Robot Policies}
Diffusion and flow models---which we treat as interchangeable here, following~\citet{gao2025diffusion}---have become a prominent policy class in robot learning because they model complex, high-dimensional continuous distributions with a simple, scalable training objective.
Diffusion Policy first showed that denoising models can serve as effective visuomotor policies for manipulation~\citep{chi2023diffusionpolicy}, and the idea has since scaled into generalist regimes~\citep{octo2024,liu2024rdt,nvidia2025groot,tri2025lbm} and flow-matching VLAs that append a generative action head to a vision-language backbone~\citep{black2024pi0,intelligence2025pi05}.
Generative policies are also beginning to appear in locomotion and whole-body control: BeyondMimic trains a latent diffusion policy for humanoid motion imitation with classifier guidance at test time~\citep{liao2025beyondmimic}, OmniXtreme pretrains a flow-matching motion prior~\citep{wang2026omnixtreme}, and PDP frames physics-based character animation as a diffusion-policy imitation problem~\citep{truong2024pdp}.

Much of the success above is based on supervised learning, and optimizing these generative policies against environment rewards remains an open problem.
Existing reinforcement-learning approaches have taken two main forms. One line freezes the generative prior and learns a small RL-trained policy on top---a Gaussian PPO residual~\citep{wang2026omnixtreme}, a latent-noise steering policy~\citep{wagenmaker2025steering}, an action-editing token over a frozen VLA~\citep{xu2026rl}, or a residual policy whose collected data is distilled back into the base VLA~\citep{xiao2025self}.
Another line trains the generative policy directly with on-policy policy gradients: FPO and FPO++ replace the log-likelihood in PPO's ratio with the conditional flow-matching loss~\citep{mcallister2025fpo,yi2026fpopp}, and ReinFlow and $\pi_{\mathrm{RL}}$ inject noise into the flow sampler to obtain tractable likelihoods~\citep{zhang2025reinflow,chen2025pirl}. However, the on-policy nature of these algorithms makes them rather sample inefficient.
Instead, we propose an online off-policy RL algorithm that directly trains flow policies from replay, whether learning from scratch or fine-tuning pre-trained flow policies.

\subsection{Off-Policy RL with Diffusion and Flow Actors}
Diffusion and flow actors have been extensively studied on simulated
continuous-control benchmarks in both offline and online
RL~\citep{wang2023diffusionql,yang2023dipo,lv2025flowbased}.
In offline RL, expressive generative models support value-based policy
improvement while anchoring the learned policy to a fixed
dataset~\citep{wang2023diffusionql,park2025fql}.
Our focus is online off-policy learning, where the actor and critic
are continually updated from an evolving replay buffer.

A central design choice is how to translate critic information into
updates of a generative policy. Some methods differentiate through
the denoising process~\citep{wang2023diffusionql,wang2024dacer,zhang2025sacflow},
while others avoid sampler backpropagation using one-step actors
or supervised objectives~\citep{park2025fql,ding2024qvpo,
ma2025dpmdsdac,psenka2023qsm,yang2023dipo,li2026qam,li2026rfm}.
DIPO~\cite{yang2023dipo} and FlowDPG~\cite{shi2026flowdpg} use critic
gradients to construct improved action or velocity regression
targets.
Recent work OGPO~\cite{patil2026ogpo} and RL-100~\cite{lei2025rl100} instead use off-policy critics in a PPO-style objective over the generative process, clipping the likelihood ratio of
every denoising step as in PPO.
\algname{} instead follows the critic's action gradient $\nabla_a Q$, and maximizes $Q$ through a one-step prediction
constructed from a replay action, together with standard flow
matching. It clips in velocity space, rather than on a likelihood ratio, to limit the actions at which the critic is queried.
This yields a simple replay-based actor update that
requires no sampler backpropagation and fits directly into a
TD3-style training loop.

The closest work, FlowRL~\citep{lv2025flowbased}, similarly combines
$Q$-maximization with flow matching, but stabilizes learning by
applying $\tanh$ to the flow endpoint, creating a mismatch between
the flow-matching target and executed action; for pretrained
policies, the added squashing also changes the action mapping at
initialization, so fine-tuning no longer starts from the pretrained
behavior. \algname{} instead clips
the $Q$-induced correction in velocity space around the
replay-conditioned flow target, constraining critic-driven updates
near replay-supported actions while retaining standard flow matching.

\subsection{Stabilizing Off-Policy Q-Learning}
A core challenge in off-policy Q-learning is that actor updates can chase critic overestimates, pulling the policy toward actions where the critic is unreliable. Classical actor-critic methods mitigate this with replay, target networks, and entropy regularization~\citep{lillicrap2016continuous,fujimoto2018td3,haarnoja2018sac}, while offline RL methods add explicit support constraints~\citep{fujimoto2019bcq,kumar2019bear,mao2023str}, critic-side conservatism~\citep{kumar2020cql}, or behavior-regularized actor extraction~\citep{fujimoto2021td3bc,kostrikov2022iql}; recent systems works make Q-learning stable at scale through wider normalization-heavy critics, target-network removal via BatchRenorm, distributional value estimates, and large-batch GPU-parallel training~\citep{li2023pql,nauman2024bro,lee2025simba,bhatt2024crossq,seo2025fasttd3}.
\algname{} integrates directly into TD3-style off-policy training. Its critic is trained as in TD3, requiring only actions sampled from the flow and no log-likelihoods or entropy estimates, and its actor update takes the form of a TD3 actor step, backpropagating $\nabla_a Q$ into the policy through a one-step flow prediction. It therefore can leverage standard stabilization techniques and recent advances for stable Q-learning at scale.

\suppressfloats[t]
\begin{figure}[t]
    \centering
    \begin{subfigure}[b]{0.48\linewidth}
        \centering
        \includegraphics[width=\linewidth]{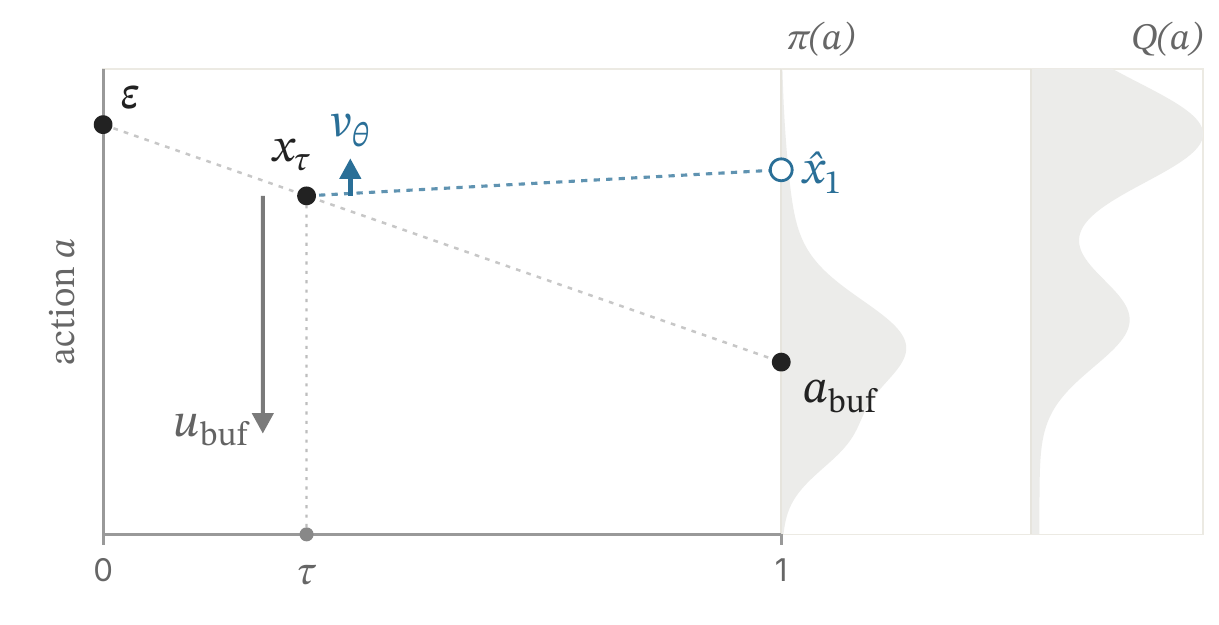}
        \caption{One-step approximation $\hat{x}_1$}
        \label{fig:algorithm-predict}
    \end{subfigure}\hfill
    \begin{subfigure}[b]{0.48\linewidth}
        \centering
        \includegraphics[width=\linewidth]{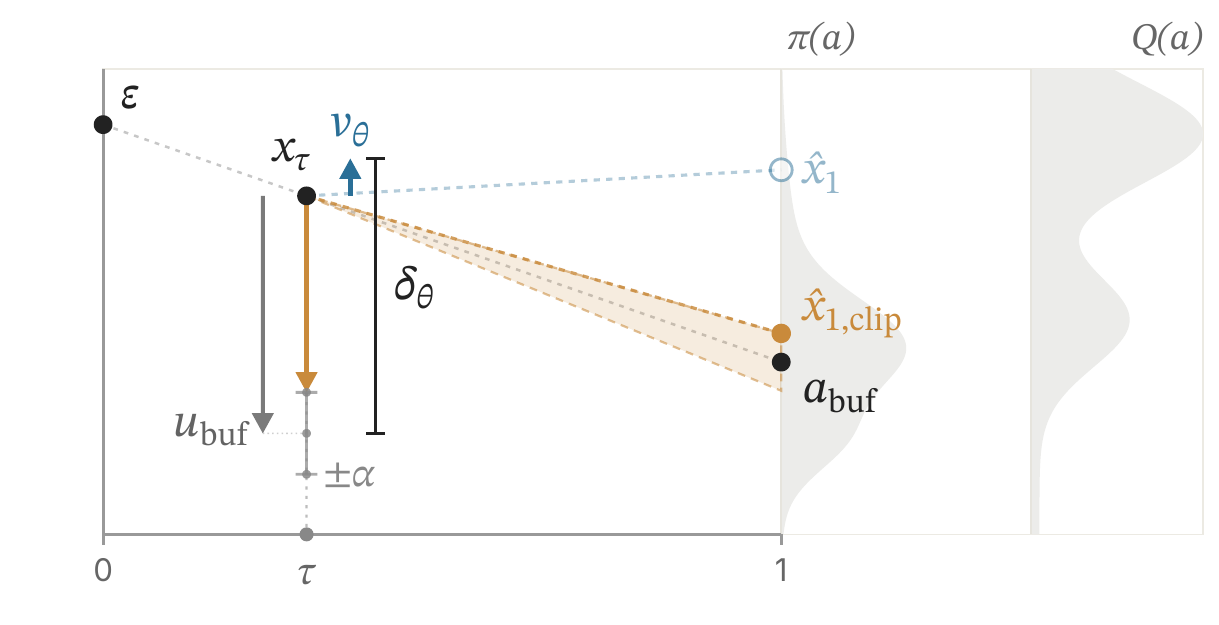}
        \caption{Clipped reconstruction $\hat{x}_{1,\mathrm{clip}}$}
        \label{fig:algorithm-geometry}
    \end{subfigure}\\[6pt]
    \begin{subfigure}[b]{0.48\linewidth}
        \centering
        \includegraphics[width=\linewidth]{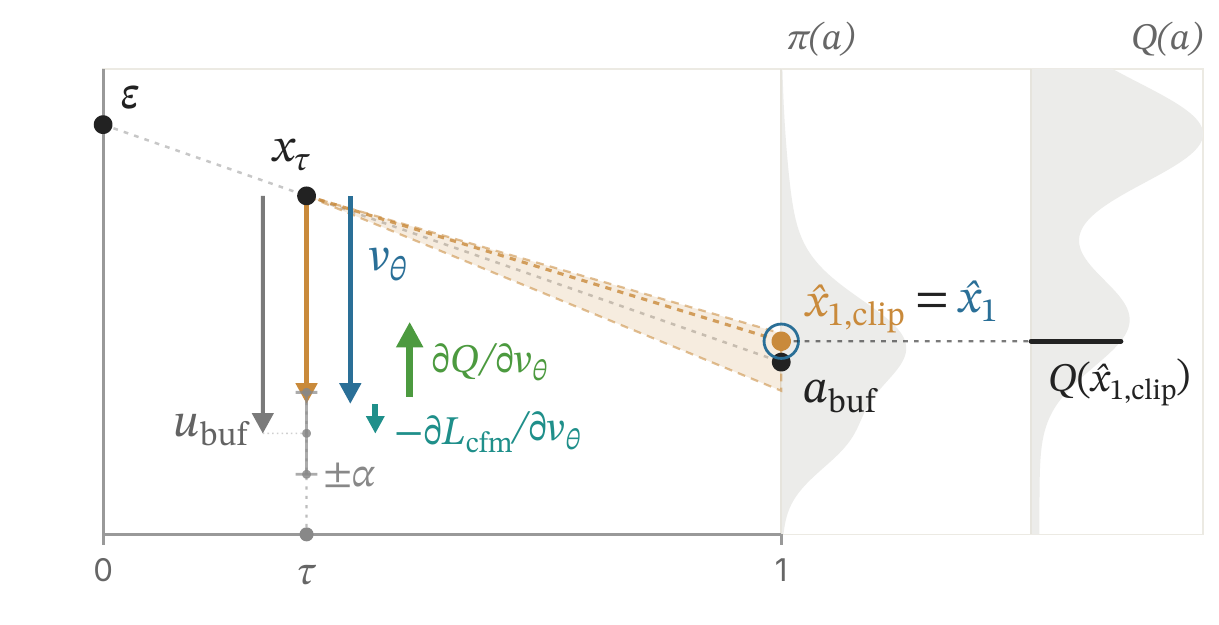}
        \caption{Inside the trust region: $\hat{x}_{1,\mathrm{clip}} = \hat{x}_1$}
        \label{fig:algorithm-inside}
    \end{subfigure}\hfill
    \begin{subfigure}[b]{0.48\linewidth}
        \centering
        \includegraphics[width=\linewidth]{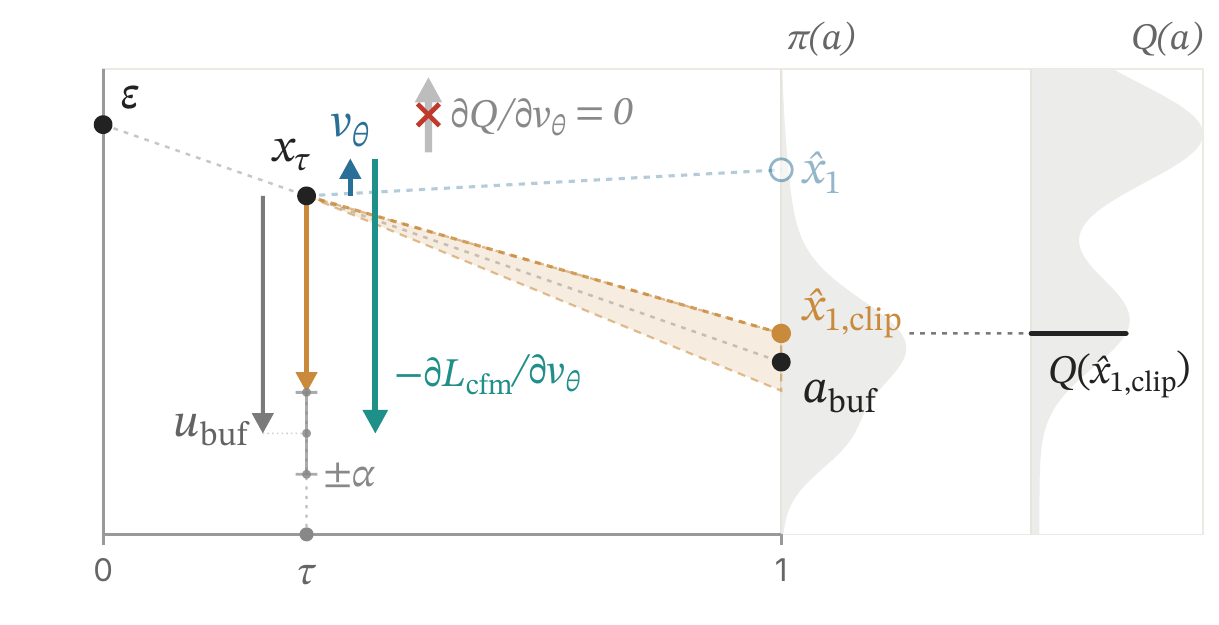}
        \caption{Outside the trust region: $\hat{x}_{1,\mathrm{clip}} \neq \hat{x}_1$}
        \label{fig:algorithm-loss}
    \end{subfigure}
    \caption{\textbf{The \algname{} actor update in 1D.}
    A one-step prediction from a noised replay action can land far from the replay action, where the critic's estimates are unreliable (a).
    Clipping the velocity keeps
    the critic's query near the replay action (b). Inside this trust region,
    the critic gradient improves the policy (c); outside it, the critic gradient is dropped and only the flow-matching loss on the replay action acts, pulling the policy back toward the replay action (d).}
    \label{fig:algorithm-overview}
\end{figure}

\section{Algorithm}
\label{sec:algorithm}

{We introduce \algname{}, an online off-policy algorithm for training flow-based actors. \algname{} uses standard TD3-style techniques for storing transitions $(s,a,r,s')$ in a replay buffer $\mathcal{D}$ and for training critics with Bellman backups, but represents the actor with a velocity field $v_\theta(s,x_\tau,\tau)$. At evaluation time, actions are sampled by drawing $x_0=\epsilon$, with $\epsilon \sim \mathcal{N}(0,I)$, and integrating $dx_\tau/d\tau = v_\theta(s,x_\tau,\tau)$ from $\tau=0$ to $\tau=1$ using $N$ Euler steps.}

\algname{} aims to steer a flow actor toward higher-value actions with the critic gradient $\nabla_a Q$, which cheaply provides an improvement direction in each action dimension. However, the exact gradient of $Q$ with respect to the flow parameters runs through every integration step of the sampler, and backpropagating through all of them is costly and can be unstable~\citep{wang2023diffusionql,wang2024dacer,zhang2025sacflow,park2025fql}. We therefore approximate the final action with a single Euler step from a noised replay action and maximize $Q$ at that prediction, while flow matching on replay actions keeps the policy close to the data. Because this update can push the prediction away from the actions that the critic is trained on,
we clip the prediction in velocity space so that the critic is queried only near replay actions. We first present the actor objective, then explain how this velocity-space trust region stabilizes policy improvement.

\subsection{\algname{} Actor Objective}
The \algname{} actor objective consists of two terms: a conditional flow-matching (CFM) term that anchors the actor to replay-buffer actions and a critic term for policy improvement.

\subsubsection{Anchoring the flow with the replay buffer.}
We construct the CFM term from a replay transition $(s,a_{\mathrm{buf}},r,s')$ by sampling
$\epsilon \sim \mathcal{N}(0,I)$ and $\tau \sim \mathcal{U}(0,1)$ and forming
\[
    x_\tau = (1-\tau)\epsilon + \tau a_{\mathrm{buf}},
    \qquad
    u_{\mathrm{buf}} = a_{\mathrm{buf}}-\epsilon
\]
where $u_{\mathrm{buf}}$ is the conditional velocity that transports $\epsilon$ to $a_{\mathrm{buf}}$ along the linear path.
We supervise the predicted velocity with the standard CFM loss:
\begin{equation}
    \mathcal{L}_{\mathrm{cfm}}
    =
    \left\|
        v_\theta(s,x_\tau,\tau)-u_{\mathrm{buf}}
    \right\|^2
    \label{eq:actor-cfm-loss}
\end{equation}
On its own, this flow-matching loss trains the actor to reproduce the replay distribution, that is, it regularizes the actor toward the behavior policy that collected the data, which the replay buffer approximates non-parametrically, as in behavior-regularized actor updates~\citep{fujimoto2021td3bc}.
By regularizing towards this behavior policy, we also keep the actor close to the actions for which the critic's estimates are most reliable.
The policy improves over time as the actor uses the critic gradients to steer its actions towards higher-value actions, and learns to flow towards these improved actions, which are placed into an evolving replay buffer.

\subsubsection{Critic-driven policy improvement.}

The core contribution of \algname{} is how critic gradients reach the velocity field. Directly maximizing the critic $Q$ through a one-step prediction of the action is simple and cheap, but can be unstable.
This instability does not come from the sampler, since the update never backpropagates through it, but from where the critic is queried. The critic is evaluated at the one-step prediction, which can drift away from replay actions, and the critic's estimates can become unreliable.
Integrating the flow ODE from $x_\tau$ with a single Euler step gives a one-step approximation of the action the policy would produce, and we maximize the critic through it:
\begin{equation}
    \begin{aligned}
    \hat x_1 = x_\tau + (1-\tau)v_\theta(s,x_\tau,\tau)
    ,\\
    \mathcal L_{\rm direct}
    = -Q(s,\hat x_1) + \lambda_{\rm cfm}\mathcal L_{\rm cfm}.
    \end{aligned}
\end{equation}
Like a TD3 actor, this update backpropagates $\nabla_a Q$ into the policy, but through a single velocity evaluation. The one-step approximation is exact when $v_\theta = u_{\mathrm{buf}}$, which gives $\hat x_1 = a_{\mathrm{buf}}$. However, in online training, a randomly initialized actor can push $\hat x_1$ off the critic's support (Fig.~\ref{fig:algorithm-predict}), where it can chase overestimated $Q$-values and the velocity magnitudes diverge.

{To address this instability, we introduce a clipped actor objective
that induces a local trust region in velocity space. By clipping
the policy's velocity around the replay-conditioned flow target,
we keep the action evaluated by the critic close to the replay action, where both the one-step approximation and the critic remain valid}.
Let $\delta_\theta := v_\theta(s,x_\tau,\tau) - u_{\mathrm{buf}}$ be the policy-buffer gap, the difference between the predicted and replay-conditioned velocities.
We clip it coordinate-wise to $[-\alpha,\alpha]$ and evaluate the critic at the corresponding one-step action:
\begin{equation}
    \begin{aligned}
    \hat{v}_{\theta,\mathrm{clip}}
    &=
    u_{\mathrm{buf}}
    +
    \operatorname{clip}(\delta_\theta,-\alpha,+\alpha),
    \\
    \hat{x}_{1,\mathrm{clip}}
    &=
    x_\tau + (1-\tau)\hat{v}_{\theta,\mathrm{clip}}
    \\
    &=
    a_{\mathrm{buf}} + (1-\tau)\operatorname{clip}(\delta_\theta,-\alpha,+\alpha),
    \end{aligned}
    \label{eq:clipped-reconstruction}
\end{equation}
where the last equality uses $x_\tau + (1-\tau)u_{\mathrm{buf}} = a_{\mathrm{buf}}$, so the critic is evaluated only within $(1-\tau)\alpha$ of the replay action in each coordinate (shaded wedge in Fig.~\ref{fig:algorithm-geometry}).
Finally, \algname{} minimizes
\begin{equation}
    \mathcal{L}_{\mathrm{actor}}
    :=
    -Q(s,\hat{x}_{1,\mathrm{clip}})
    +
    \lambda_{\mathrm{cfm}} \mathcal{L}_{\mathrm{cfm}}.
    \label{eq:actor-loss}
\end{equation}
This separation is important: when the policy-buffer gap is large, clipping prevents the resulting endpoint from moving arbitrarily far from the replay action in the $Q$ term, while the flow-matching loss still penalizes the full, unclipped gap.

\subsection{Filtered Q-Gradients as a Trust Region}
Clipping not only bounds the one-step reconstruction, but also
filters the critic gradients that reach the actor.
Differentiating the $Q$ term in Eq.~\ref{eq:actor-loss}
{through the clipped
reconstruction in Eq.~\ref{eq:clipped-reconstruction} gives}
\begin{equation}
    \begin{aligned}
    \frac{\partial Q}{\partial v_\theta}
    &=
    (1-\tau)\,
    m \odot
    \nabla_a Q(s,a)\big|_{a=\hat{x}_{1,\mathrm{clip}}},
    \\
    m_i &=
    \mathbbm{1}\!\left[|\delta_{\theta,i}| < \alpha\right]
    \end{aligned}
    \label{eq:q-gradient-mask}
\end{equation}
{For each replay sample, velocity component $i$ receives a $Q$-gradient only when its policy-buffer gap lies within the
trust-region bounds (Fig.~\ref{fig:algorithm-inside}). When that component saturates the clip, its $Q$-gradient is masked out, while the CFM term remains active and pulls the raw velocity prediction toward $u_{\mathrm{buf}}$ (Fig.~\ref{fig:algorithm-loss}). If the gap re-enters the clipping range, the $Q$-gradient resumes.}

{The trust region thus acts as a policy-consistency filter: critic gradients drive local improvement within the allowed velocity range, while flow matching pulls predictions outside that range back toward the replay-conditioned target.}
This is analogous to the ratio clip of PPO~\cite{schulman2017ppo}, which likewise enforces a trust region by withholding the policy-improvement gradient.
We write an interpretation of this analogy through the CFM--ELBO correspondence of Kingma and Gao~\cite{kingma2023understandingdiffusionobjectiveselbo}, reading the clip as a trust region on an ELBO surrogate of the replay action, and sketch this out in Appendix~\ref{supp:clip-elbo}.

\newif\ifqfOneColumn
\ifdim\columnwidth>0.7\textwidth\qfOneColumntrue\fi

\section{Analyzing Filtered Q-Gradients}
\label{sec:analysis}

\subsection{Analyzing the velocity clip}
\begin{figure}[!t]
    \centering
    \newlength{\qfClipBoxH}\setlength{\qfClipBoxH}{2.2cm}%
    \newlength{\qfClipSchemH}\setlength{\qfClipSchemH}{\dimexpr\qfClipBoxH*1000/741\relax}%
    \newcommand{\qfClipPanel}[1]{\raisebox{0pt}[\dimexpr\qfClipBoxH+0.55cm\relax][0.8cm]{#1}}%
    \tikzset{every picture/.append style={baseline=(current axis.south)}}%
    \begin{subfigure}[b]{0.27\linewidth}
        \centering
        \pgfplotsset{flop clamp curves main/.append style={scale only axis, width=3.1cm, height=\qfClipBoxH, title style={font=\sffamily\scriptsize, yshift=0.85pt}, legend style={font=\sffamily\tiny, nodes={font=\sffamily\tiny}}},
            flop clamp main legend/.style={legend style={at={(axis description cs:0.03,0.97)}, anchor=north west}}}%
        \expandafter\def\csname flopAlgoLabel@FQR-noclamp\endcsname{Direct Q}%
        \qfClipPanel{\flopClampAblationMain{figures/data/mujoco/clamp_ablation/tidy}}
        \caption{Unclipped collapse.}
        \label{fig:noclamp-ablation-main}
    \end{subfigure}\hfill
    \begin{subfigure}[b]{0.36\linewidth}
        \centering
        \qfClipPanel{\raisebox{-0.159\qfClipSchemH}{\includegraphics[height=\qfClipSchemH]{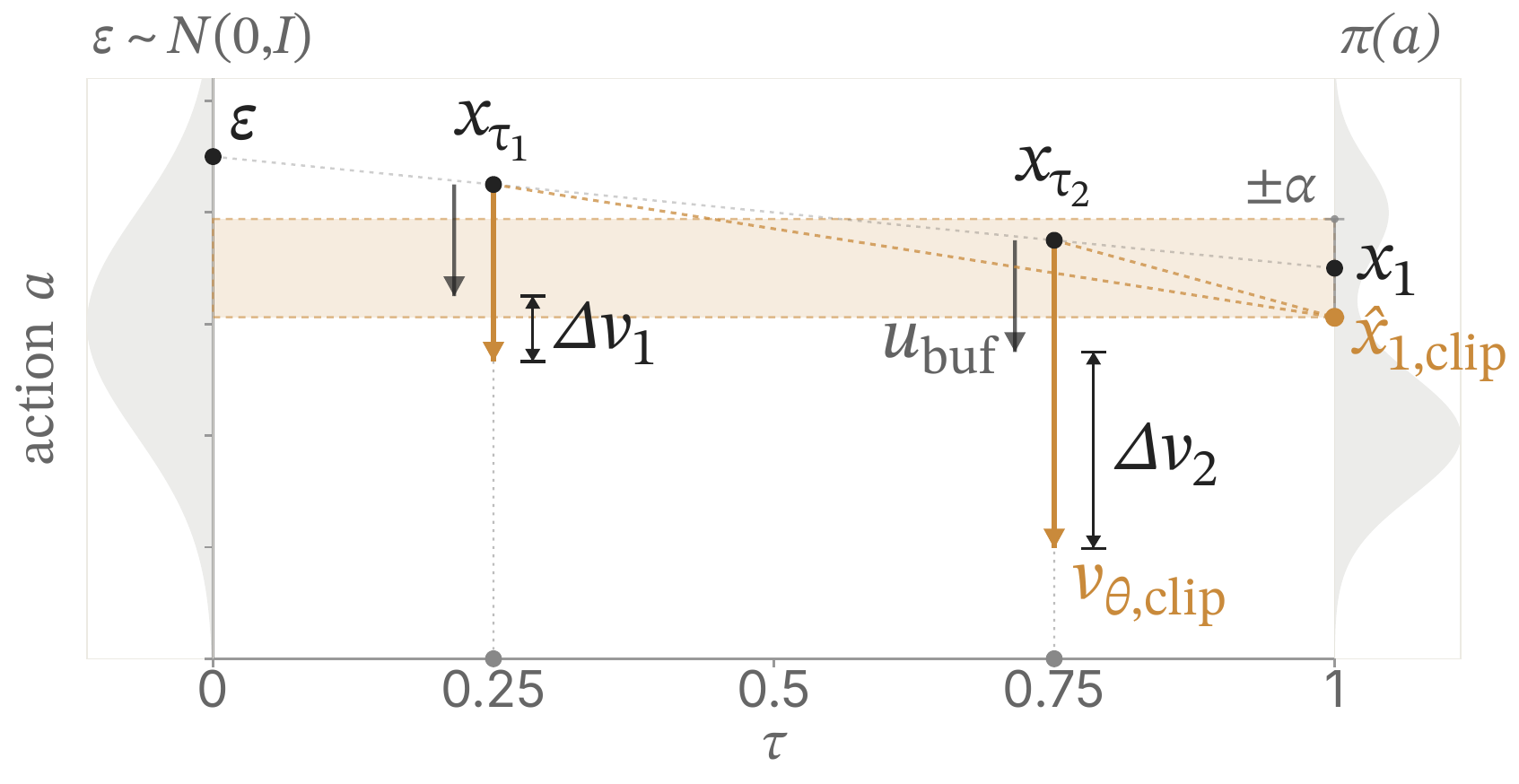}}}
        \caption{Action-space clamp geometry.}
        \label{fig:action-clamp-schematic}
    \end{subfigure}\hfill
    \begin{subfigure}[b]{0.33\linewidth}
        \centering
        \pgfplotsset{flop flowfield pair/.append style={title style={font=\sffamily\scriptsize, yshift=-2.8pt}}}%
        \qfClipPanel{\flopFlowfieldClampRow{figures/data/mujoco/clamp_flowfield/tidy}{1.6cm}{\qfClipBoxH}}
        \caption{Learned flow fields.}
        \label{fig:flowfield-clamp-pair}
    \end{subfigure}
    \caption{\textbf{Analyzing the velocity clip.}
    (a)~Without the clip, Direct Q collapses on Humanoid-v4 even at a $10\times$ lower actor learning rate.
    (b)~Clamping the critic input to a fixed band around $a_\mathrm{buf}$ instead allows a policy-buffer gap of up to $\alpha/(1-\tau)$, without bound as $\tau \to 1$.
    (c)~On HalfCheetah-v4, the learned flow field then bends sharply near $\tau = 1$ (circled), which the velocity clip avoids.}
    \label{fig:clipping-compare}
\end{figure}

{In this section,} we analyze the \algname{}'s actor update described in Sec.~\ref{sec:algorithm} within Gymnasium~\cite{towers2026gymnasium} MuJoCo-v4.
Results are plotted in Figure~\ref{fig:clipping-compare}.
{We } find that:

\textbf{The velocity-space clipping in the actor loss is essential for stable training.} The clip in \algname{}'s actor update (Eq.~\ref{eq:clipped-reconstruction}) prevents large, destabilizing updates from the critic early in training when the value estimates are unreliable, which in turn spikes the magnitude across the velocity field, debilitating learning. We verify this claim on Humanoid-v4 against \emph{Direct Q (unclamped)}, an ablation that applies the critic's gradient to the actor without the per-element velocity clip (Fig.~\ref{fig:noclamp-ablation-main}), and find that the unclamped update collapses learning even at a $10\times$ lower actor learning rate (\mbox{3e-5} instead of the default \mbox{3e-4}), while the default \algname{} clip recovers a usable policy. A similar collapse seems to reappear when finetuning a pretrained manipulation policy (Sec.~\ref{sec:exp:manipulation}), where the unclamped update drops a policy that already solves the task to near-zero success within the first $100$k steps before recovering.

\textbf{Velocity-space clipping creates a lower-curvature flow field than action-space clipping.} The velocity-space clip in \algname{} effectively induces a $\tau$-dependent action-space trust region of size $(1-\tau)\alpha$ (see Fig.~\ref{fig:algorithm-geometry}).
Here, we ablate this decision by clipping the correction from {$\hat{x}_1$}
instead, effectively making $\alpha$ time-dependent ($\frac{\alpha}{1-\tau}$; Fig.~\ref{fig:action-clamp-schematic}).
We find that although this may lead to a similar policy performance, the learned flow field may have a much higher curvature (Fig.~\ref{fig:flowfield-clamp-pair}). The reason is that a fixed action-space bound $\alpha$ translates to a velocity bound of $\alpha/(1-\tau)$, which is unbounded as $\tau \to 1$: the critic can inject arbitrarily large velocity corrections exactly where the field determines the final action, and the learned field bends sharply to accommodate them.

{\textbf{The velocity-space clip remains active throughout training.}} Since the \textit{Direct Q (unclamped)} collapse occurs early in training, one may expect the clip to have no effect once the critic is reliable and the policy is stable. We test this by running \algname{} on Ant and Humanoid with $\alpha=\frac{1}{2}$, and logging the fraction of action dimensions whose correction lies within the clip, $|\delta_{\theta,i}| < \alpha$, and hence receives a $Q$-gradient (Eq.~\ref{eq:q-gradient-mask}). We find that the clipped minority shrinks as the critic improves but does not vanish, so the clip continues to filter critic corrections even after the policy is performant: at $50$k, $250$k, and $500$k steps, this fraction is $74/83/85\%$ on Ant and $81/83/82\%$ on Humanoid. This suggests that the clip is therefore not only a safeguard against an unreliable early critic, but may act as a trust region that constrains the actor update for the entirety of training.

\subsection{Validation on MuJoCo-v4}
\label{sec:exp:mujoco}

We use the same Gymnasium MuJoCo-v4 environments to compare \algname{} against recent off-policy flow-based RL methods, where all share the same TD3 harness, flow actor, and critic (Appendix~\ref{supp:mujoco}). This comparison isolates how they inject critic information into the flow actor: (i) as a \emph{regression target}, where QSM~\citep{psenka2023qsm} directly aligns $v_\theta$ with $\nabla_a Q$; (ii) as a \emph{regularized objective}, where FlowRL~\citep{lv2025flowbased} pairs $Q$-maximization with a Wasserstein-2 penalty; and (iii) as \emph{improved targets}, where DIPO~\citep{yang2023dipo} improves replay actions under the critic and refits the actor to them.
\algname{} instead uses the pathwise Q gradient $\nabla_a Q$ to update the actor.

\begin{wrapfigure}{r}{0.52\textwidth}
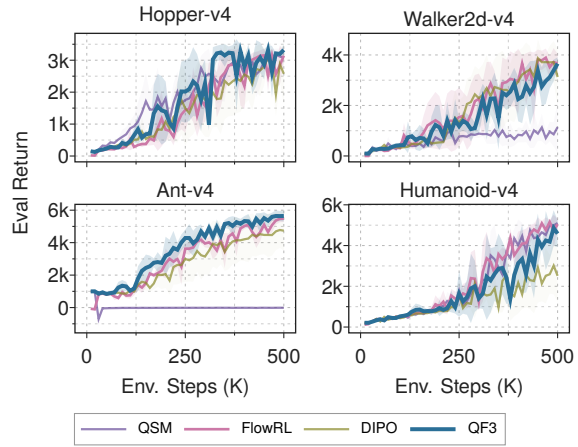

    \vspace{-12pt}
    \centering
    \pgfplotsset{flop mujoco two by two/.style={height=1.7cm, title style={font=\sffamily\scriptsize}}}%
    \pgfplotsset{flop mujoco two by two legend/.style={legend style={column sep=3pt, inner xsep=2pt, font=\sffamily\tiny, nodes={font=\sffamily\tiny}}}}%
    \flopMujocoAllCurvesTwoByTwo{figures/data/mujoco/tidy}
    \vspace{2pt}
    \captionsetup{justification=raggedright,singlelinecheck=false,skip=3pt,font=small}%
    \caption{\textbf{\algname{} in MuJoCo-v4~\citep{towers2026gymnasium}}.
    The plots show evaluation returns over $500$k environment steps (mean $\pm$ std across $3$ seeds).
    \algname{} performs competitively with state-of-the-art flow RL methods.
    }
    \label{fig:mujoco-curves}
    \vspace{-12pt}
\end{wrapfigure}
We evaluate on Hopper, Walker2d, Ant, and Humanoid (Fig.~\ref{fig:mujoco-curves}). Every flow method is tuned with the same protocol: we sweep its method-specific hyperparameters (Table~\ref{tab:mujoco-hyperparameters}) and select the best configuration per task by mean return over steps $450$k--$500$k and seeds. \algname{} performs competitively with one velocity evaluation per actor update, whereas DIPO takes \mbox{$10$--$40$} action-gradient steps and FlowRL trains an extra expectile value function. QSM fails on Ant and lags far behind on Walker2d for every score-loss weight in its sweep (Fig.~\ref{fig:supp-qsm-hp-sweep}), suggesting that matching the gradient's scale is harder than following its direction. A single \algname{} setting ($\alpha=1$, $\lambda_{\mathrm{cfm}}=0.3$) reaches at least $79\%$ of its per-task best return on all four tasks (Fig.~\ref{fig:supp-qf3-velocity-sweep}).

\section{Experimental Results}
\label{sec:result}

We evaluate \algname{} in the two regimes:
first, we scale to humanoid control, combining the actor update with the high-throughput training recipe of FastTD3~\cite{seo2025fasttd3}. This allows \algname{} to train velocity- and motion-tracking policies in Holosoma~\citep{amazonfar2025holosoma} at wall-clock speeds much faster than the on-policy flow baseline FPO++~\citep{yi2026fpopp}---the first off-policy flow RL method, to our knowledge, to train a humanoid policy from scratch and transfer zero-shot to hardware (Sec.~\ref{sec:exp:locomotion}, Fig.~\ref{fig:humanoid-sim2real-curves}).
Second, we fine-tune pretrained manipulation policies in ABC Sim, where \algname{} trains LoRA adapters on a VLA's flow head (Sec.~\ref{sec:exp:manipulation}), and on Robomimic, where it updates the full policy (Sec.~\ref{sec:exp:robomimic}).

\subsection{Evaluating \algname{} for humanoid locomotion}
\label{sec:exp:locomotion}

\begin{figure}[!t]
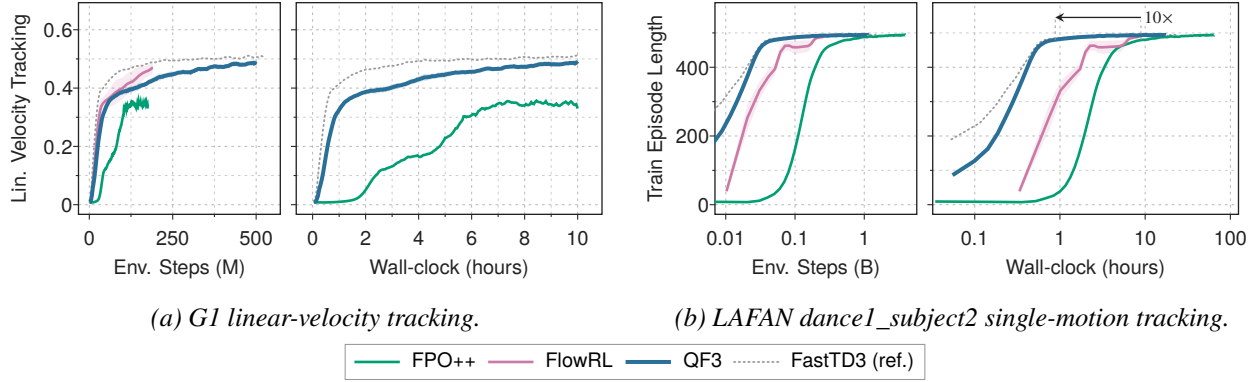

    \centering
    \flopLocomotionResultsOneRow{figures/data/locomotion/tidy/smoothed}
    \caption{\textbf{Humanoid locomotion learning curves.}
    Each panel shows env.-step (left) and wall-clock (right) views of the same task.
    (\subref{fig:splash-vel-curves})~Per-step \texttt{raw\_lin\_vel} reward under a fixed 10-hour budget on a single A10G.
    (\subref{fig:splash-mot-curves})~Mean episode length (max 500) on LAFAN dance motion, in log scale. \algname{} converges $\approx 10\times$ faster in wall-clock than FPO++~\citep{yi2026fpopp}. FlowRL's velocity-tracking runs end at 190M env steps.
    }
    \label{fig:locomotion-curves}
\end{figure}

\newcommand{\qfManipulationFigure}{%
\begin{figure*}[!t]
    \centering
    \input{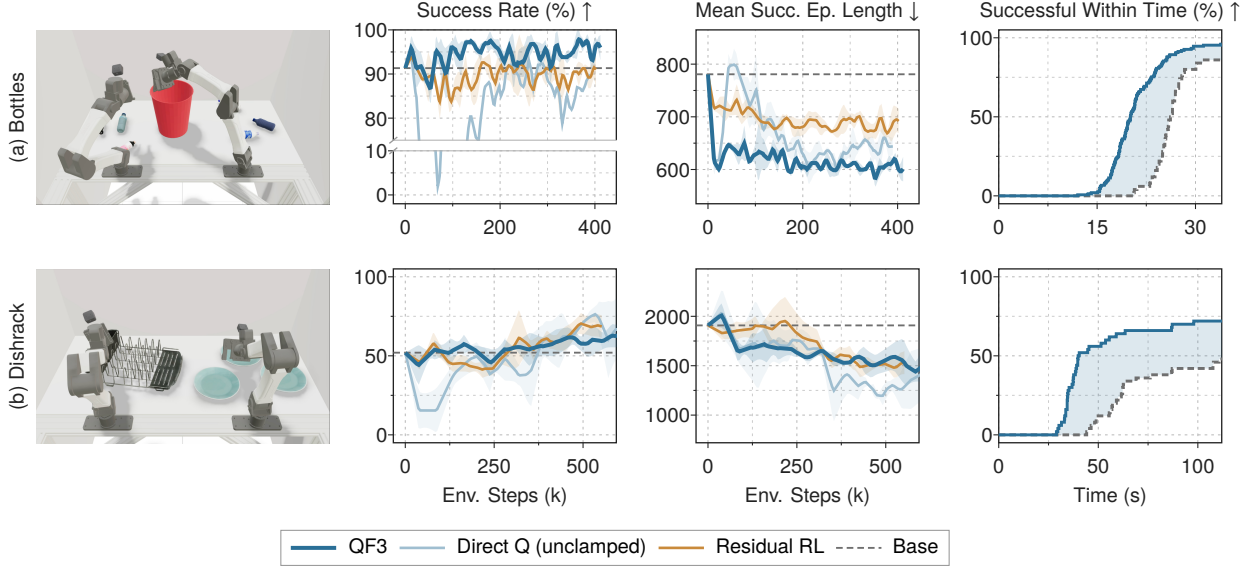}
    \caption{\textbf{ABC-VLA manipulation fine-tuning.}
    (\subref{fig:manipulation-bottles}) \emph{Throw Bottles in Bin} and (\subref{fig:manipulation-dishrack}) \emph{Load Plates in Dishrack}, 3~seeds, mean~$\pm$~std. The right column shows the share of episodes solved within each time for the pretrained base policy and the final checkpoint of one \algname{} run. The shaded gap is the drop in mean episode time, counting failures at the limit ($5.8$\,s, $27$\,s).
    }
    \label{fig:manipulation-finetuning}
\end{figure*}
}
\ifqfOneColumn\else\qfManipulationFigure\fi

Flow policies are increasingly used as pretrained priors for humanoid whole-body control~\citep{liao2025beyondmimic,wang2026omnixtreme}, but improving them with RL still relies on residual policies over a frozen prior~\citep{wang2026omnixtreme}. As a first step toward native flow RL in this setting, we ask whether \algname{} can train a flow policy from scratch within FastTD3's massively parallel, high-throughput recipe as efficiently as FastTD3 itself.

To study this, we train policies using Holosoma~\citep{amazonfar2025holosoma}, an open-source humanoid whole-body RL framework that supplies both the task suite (velocity tracking, single-motion tracking) for the 29-DoF Unitree G1 and a tuned FastTD3~\citep{seo2025fasttd3} baseline,
and compare against FPO++ by porting the mixed rough terrain environment from Holosoma into their open-source code release.
FastTD3 is a high-throughput off-policy recipe that pairs TD3's clipped double-Q critic and target networks with massively parallel GPU simulation, a large replay buffer, large-batch updates, and a distributional categorical critic~\citep{bellemare2017distributional}.
Because \algname{} keeps TD3's replay buffer, critic, and target-update structure (Sec.~\ref{sec:algorithm}) and only changes the actor, this recipe drops in unchanged: we replace the deterministic FastTD3 actor with the \algname{} flow actor and its clipped actor loss (Eq.~\ref{eq:clipped-reconstruction}), use the mean of the categorical critic as the scalar $Q(s,\hat{x}_1)$ in the actor update, and share all remaining hyperparameters with FastTD3 (Appendix~\ref{supp:holosoma}). At deployment, the flow is integrated deterministically from $\epsilon=0$ with $5$ Euler steps, following FPO++~\citep{yi2026fpopp}.

We report learning curves and wall-clock time comparisons in Fig.~\ref{fig:locomotion-curves}, with mean$\pm$std over 3 seeds; each run trains on a single GPU.\footnote{Velocity-tracking runs use an NVIDIA A10G and motion-tracking runs an NVIDIA L40S. FlowRL's velocity-tracking runs used an NVIDIA L4, so we omit FlowRL from the velocity wall-clock panel.} Results show that QF3 succeeds in both the humanoid locomotion and motion tracking environments. QF3 trains {an order of magnitude} faster than the on-policy baseline FPO++~\cite{yi2026fpopp},
and is competitive with Holosoma's optimized FastTD3 implementation.
On velocity tracking, \algname{} finishes within $\sim$4\% of FastTD3's final tracking reward under the same 10-hour budget. On motion tracking, \algname{} reaches full-length (500-step) episodes in roughly half an hour of wall-clock, close to FastTD3, while FPO++ takes $\approx 10\times$ longer in both wall-clock and environment steps. FlowRL~\citep{lv2025flowbased}, another off-policy flow method that we run on the same FastTD3 recipe, reaches comparable tracking performance on both tasks but is less consistent: its seeds vary about $3\times$ as much as \algname{}'s on velocity tracking and $9\times$ on motion tracking, perhaps due to the additional expectile value function it trains to regularize its actor. On motion tracking it also needs about $1.8\times$ as many environment steps as \algname{} to reach near-full-length (490-step) episodes.

We also find policies trained with QF3 can be successfully deployed on real robots zero-shot, with no real-world fine-tuning and the same joint-target action space and PD gains as in simulation: we find that the G1 humanoid policies trained using QF3 can follow velocity commands and track long, challenging, and dynamic motions (LAFAN ``dance1\_subject2", approx. 3 minutes; see Fig.~\ref{fig:humanoid-sim2real-curves}).

\subsection{Fine-tuning pretrained manipulation policies}
\label{sec:exp:finetuning}
\ifqfOneColumn\qfManipulationFigure\fi

We next ask whether the same core update can improve a pretrained generative policy rather than train one from scratch.
In both settings below, we add a flow-matching loss toward the frozen base policy, $\lambda_{\mathrm{base}}\|v_\theta-v_{\mathrm{base}}\|^2$, to the \algname{} actor loss, and find this anchor important for stable fine-tuning~\citep{jiang2026dipod}.

\subsubsection{ABC-Sim}
\label{sec:exp:manipulation}

We fine-tune the vision-conditioned ABC-VLA policy~\citep{abc2026} in ABC Sim on two bimanual tasks, \emph{Throw Bottles in Bin} and \emph{Load Plates in Dishrack} (Fig.~\ref{fig:manipulation-finetuning}).
The same pretrained policy is used for both tasks, and pairs a frozen VLM with a DiT action head that integrates a 5-step flow into a 30-step action chunk; \algname{} trains rank-4 LoRA adapters~\citep{hu2022lora} on the attention and MLP projections of the head, initialized to zero so that training starts exactly at the base policy.
We compare against residual RL, which freezes the policy and trains a bounded additive correction with TD3~\citep{dong2025expo}, sharing the same critic, replay buffer, simulator, and evaluation protocol, and against \textit{Direct Q (unclamped)}, which is \algname{} {without the velocity field trust-region.}
An episode succeeds when all five bottles are in the bin or all three plates are in the rack before the timeout (1000 and 3300 control steps, or $34$ and $112$\,s). \emph{Bottles} rewards $+1$ for each bottle the first time it enters the bin. \emph{Dishrack} rewards progress normalized by object count ($+1/3$ per newly racked plate) plus a $+1$ bonus on success.
Both tasks run at a $29.4$\,Hz control rate, and the policy executes the first $15$ steps of each chunk, about two policy calls per second. Training collects experience in $16$ parallel simulated worlds (Appendix~\ref{supp:manipulation}, Table~\ref{tab:abc-settings}).

\begin{figure*}[!b]
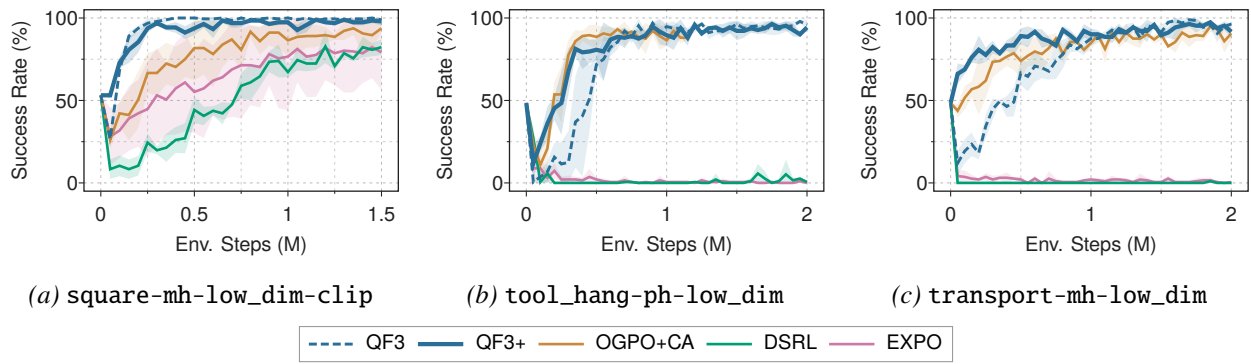

    \centering
    \flopRobomimic{figures/data/robomimic/tidy}
    \caption{\textbf{Robomimic fine-tuning.} Success rate versus online environment steps on three Robomimic tasks. All methods start from the same behavior-cloned checkpoint (step $0$) and are evaluated with the policy they act with: \algname{}, \algname{}+, and OGPO+CA with SDE sampling, DSRL with its noise policy, and EXPO with its edited actions. 3 seeds, mean $\pm$ std.}
    \label{fig:robomimic}
\end{figure*}

We report success rate and the mean length of successful
episodes from 50-episode evaluations in Fig.~\ref{fig:manipulation-finetuning}.
\algname{} improves the pretrained policy on both tasks.
On \textit{Bottles}, successful episodes finish 22\% faster (781 $\rightarrow$ 606
steps) and success rises from 91\% to 95\%; over the final 100k steps this is
12\% faster and 5 points higher than residual RL.
The speedup is spread evenly over this task: each of the five bottles takes
about 1\,s (29 steps) less.
On \textit{Dishrack},
\algname{} improves both success rate and successful-episode
length relative to the pretrained base policy, with performance
comparable to residual RL.
Here the gain comes mostly from avoiding failures, through two behavior changes (App.~\ref{supp:manipulation-diagnostics}).
On the first plate, the base policy usually grasps with the left hand, then
the right, then the left again; \algname{} instead settles on a single
right-to-left handover (which the base policy already produces occasionally),
dropping the extra grasps.
It also seats plates more upright ($9^\circ$ vs.\ $14^\circ$ median tilt),
so only 8\% of placed plates fall back out of the rack, against 21\% for the
base policy.
\textit{Direct Q (unclamped)} suffers a
pronounced transient collapse on both tasks before recovering,
consistent with the instability observed in Sec.~\ref{sec:analysis}.

\subsubsection{Robomimic}
\label{sec:exp:robomimic}

We also show full finetuning of pretrained flow policies with \algname{} on three Robomimic tasks~\citep{mandlekar2021robomimic}, \emph{Square}, \emph{Tool Hang}, and \emph{Transport} (Fig.~\ref{fig:robomimic}).
We compare against OGPO~\citep{patil2026ogpo}, which fine-tunes the full policy with PPO-style updates over its denoising steps using the critic's value as the reward, and against DSRL~\citep{wagenmaker2025steering} and EXPO~\citep{dong2025expo}, which do not fine-tune the base policy with RL but steer its input noise and edit its actions, respectively.
We follow OGPO's online-only protocol in its released codebase: all methods start from the same behavior-cloned flow policy (stopped at ${\le}50\%$ success on \emph{Square} and \emph{Tool Hang}) and use no demonstrations online.
\algname{} replaces only OGPO's actor update, and \algname{}+ adds OGPO's success-buffer regularizer, a flow-matching loss on actions from successful episodes.
For OGPO we run OGPO+CA, the variant its authors recommend, which uses the same regularizer and additionally applies conservative advantages, updating on an action only when every critic in the ensemble agrees on the sign of its advantage (Appendix~\ref{supp:robomimic}).
Following OGPO, QF3 and QF3+ act by sampling $8$ candidate actions and executing the one the critic scores highest.

Results in Figure.~\ref{fig:robomimic} show \algname{}+ matches or exceeds OGPO+CA in final success rates over the three tasks. We also find that \algname{} does not need the success-buffer regularizer that OGPO relies on: without it, \algname{} reaches $96\%$ success on \emph{Tool Hang} and $96\%$ on \emph{Transport}, whereas OGPO without it saturates near $80\%$ and $40\%$ on these tasks in its authors' experiments~\citep[Fig.~9]{patil2026ogpo}. Because \algname{}'s actor follows the critic's gradient directly, it is more sensitive to an untrained critic than OGPO's clipped policy-gradient update, so we train the critic alone on base-policy rollouts before the first actor update ($20$k steps on \emph{Square} and \emph{Tool Hang}, $100$k on \emph{Transport}). We find that DSRL and EXPO, which leave the base policy's weights unchanged, struggle to improve performance on the long-horizon tasks.

\section{Conclusion}
\label{sec:conclusion}

We present QF3, an online off-policy reinforcement-learning algorithm for flow policies. QF3 trains a flow actor directly from replay by combining standard flow matching with critic-driven improvement through a one-step prediction of the denoised action. To stabilize this update, QF3 clips critic-induced velocity corrections around the replay-conditioned flow target. This bounds where the critic is queried and masks Q-gradients when the actor deviates too far from replay, providing a simple policy-consistency filter for off-policy learning. The actor update requires neither integration nor differentiation through the flow ODE.
On continuous-control benchmarks, QF3 performs competitively with recent off-policy flow RL methods. Paired with the FastTD3 training recipe, it trains Unitree G1 velocity- and motion-tracking policies at wall-clock speeds close to FastTD3 and approximately an order of magnitude faster than FPO++, with zero-shot transfer to hardware. The same update also fine-tunes a pretrained ABC-VLA policy on two bimanual manipulation tasks in simulation, improving success rate and successful-episode duration over the base policy. On three Robomimic tasks, it finetunes the full policy to over 95\% success. Together, these results suggest that a single replay-based flow actor update can support both learning robot behaviors from scratch and improving flow policies acquired from demonstrations.

\section*{Acknowledgments}
This project was funded in part by NSF:CNS-2235013, IARPA DOI/IBC
No.~140D0423C0035, DARPA No.~HR001123C0021, BAIR Commons sponsors, and the
Bakar Fellows Program. Chung Min Kim and Brent Yi are supported by the NSF
Graduate Research Fellowship Program, Grant DGE 2146752.
We thank Qiyang Li, Zhiyuan Zhou, and Arthur Allshire for fruitful
technical conversations.

\bibliography{example,icra_additions}

\clearpage
\appendix
\setcounter{section}{0}
\renewcommand{\thesection}{A.\arabic{section}}
\setcounter{figure}{0}\setcounter{table}{0}\setcounter{equation}{0}
\renewcommand{\thefigure}{A.\arabic{figure}}
\renewcommand{\thetable}{A.\arabic{table}}
\renewcommand{\theequation}{A.\arabic{equation}}
\renewcommand{\theHfigure}{A.\arabic{figure}}
\renewcommand{\theHtable}{A.\arabic{table}}
\renewcommand{\theHequation}{A.\arabic{equation}}
\renewcommand{\theHsection}{A.\arabic{section}}

\begin{center}\LARGE\bfseries Appendix\end{center}

\section{\algname{} Implementation Details for Training from Scratch}
\label{supp:qf3}

This section describes the flow actor and critic used when training from scratch, in MuJoCo-v4 and humanoid locomotion. The fine-tuning experiments keep the pretrained policy's architecture, described in Appendices~\ref{supp:manipulation}, \ref{supp:robomimic}, and~\ref{supp:t2i}.

\paragraph{Flow actor and integration.}
The actor is an MLP that takes the observation, noisy action, and time embedding $[s,\,x_\tau,\,\phi(\tau)]$ and outputs the velocity $v_\theta(s,x_\tau,\tau)$. The flow time is encoded by a fixed sinusoidal embedding $\phi(\tau)\in\mathbb{R}^{d}$ with $d=64$, half-bank size $H=d/2$, and log-spaced frequencies $\omega_k=\exp\!\bigl(-k\log(10000)/(H-1)\bigr)$ for $k=0,\dots,H-1$, with $\phi(\tau)=[\sin(\tau\omega_k),\cos(\tau\omega_k)]_k$ stored as a non-trainable buffer. The last layer of the velocity network is initialized to zero, so the initial flow is the identity transport $x_1=\epsilon$. Rollouts use explicit forward Euler with uniform step $\Delta\tau=1/N$, $N=5$: starting from $x^{(0)}=\epsilon\sim\mathcal{N}(0,I)$, we iterate $x^{(k+1)} = x^{(k)}+\Delta\tau\,v_\theta(s,x^{(k)},k\Delta\tau)$ and take $x_1 = x^{(N)}$ as the normalized action.

Table~\ref{tab:qf3-arch} summarizes the actor and critic architectures for the MuJoCo-v4 and humanoid locomotion experiments. The two settings differ in network width, layer normalization, and critic head, but share the flow parameterization, time embedding, activation, and Euler-step count.

\paragraph{Critic parameterization.}
In MuJoCo-v4, the critic is a standard TD3 twin-Q with target smoothing and delayed actor updates, identical to the CleanRL TD3 baseline.
In the locomotion setting (Holosoma), we adopt the FastTD3 distributional critic: each Q-network outputs categorical logits over $K=101$ atoms uniformly spaced on $[-V_{\max}, V_{\max}]$ with $V_{\max}=20$, trained with a cross-entropy loss against a projected target distribution. The mean of the predicted distribution is the scalar Q-estimate used by the actor update.

\begin{table}[!h]
    \centering
    \footnotesize
    \begin{tabularx}{\linewidth}{@{}>{\raggedright\arraybackslash}X
                                    >{\raggedright\arraybackslash}p{0.22\linewidth}
                                    >{\raggedright\arraybackslash}p{0.30\linewidth}@{}}
        \toprule
        & MuJoCo-v4 & Locomotion \\
        \midrule
        Actor MLP hidden width & $256$ & $512$ \\
        Actor MLP hidden layers & $2$ & $2$ \\
        Actor activation & SiLU & SiLU \\
        Actor LayerNorm & no & yes \\
        Time embedding dim. & $64$ & $64$ \\
        Euler steps $N$ & $5$ & $5$ \\
        \addlinespace
        Critic head & scalar twin-Q (TD3) & distributional twin-Q (FastTD3) \\
        Critic MLP hidden width & $256$ & $768$ \\
        Critic MLP hidden layers & $2$ & $2$ \\
        Critic atoms / support & --- & $101$ over $[-20, 20]$ \\
        \bottomrule
    \end{tabularx}
    \caption{\textbf{\algname{} actor and critic architectures.} MuJoCo-v4 uses the CleanRL-based implementation (Appendix~\ref{supp:mujoco}); locomotion uses the FastTD3-based implementation for the humanoid and sim-to-real experiments (Appendix~\ref{supp:holosoma}).}
    \label{tab:qf3-arch}
\end{table}

\section{Velocity Clipping as an ELBO-Surrogate Trust Region}
\label{supp:clip-elbo}

This section interprets the velocity clip of Sec.~\ref{sec:algorithm} as a trust region on an ELBO surrogate of the replay action.
For a replay action $a_{\mathrm{buf}}$ and Gaussian noise $\epsilon$, define
\begin{equation}
    \begin{gathered}
    x_\tau=(1-\tau)\epsilon+\tau a_{\mathrm{buf}},
    \qquad
    u_{\mathrm{buf}}=a_{\mathrm{buf}}-\epsilon,
    \\
    \delta_\theta=v_\theta(s,x_\tau,\tau)-u_{\mathrm{buf}} .
    \end{gathered}
    \label{eq:supp-cfm-path}
\end{equation}
With the expectation taken over $(s,a_{\mathrm{buf}})\sim\mathcal D$, $\epsilon\sim\mathcal N(0,I)$, and $\tau\sim\mathcal U(0,1)$, the dimension-normalized CFM loss is
\begin{equation}
    \mathcal L_{\mathrm{CFM}}(\theta)
    =
    \mathbb E
    \left[
        \frac{1}{d_a}
        \|v_\theta(s,x_\tau,\tau)-u_{\mathrm{buf}}\|_2^2
    \right].
    \label{eq:supp-cfm-loss}
\end{equation}

The likelihood connection follows from \citet{kingma2023understandingdiffusionobjectiveselbo}. For noised data $z_\lambda=\alpha_\lambda x+\sigma_\lambda\epsilon$, the weighted denoising loss is
\begin{equation}
    \mathcal L_w(x)
    :=
    \frac{1}{2}
    \mathbb E_{\lambda\sim p(\lambda),\,\epsilon}
    \left[
        \frac{w(\lambda)}{p(\lambda)}
        \|\hat\epsilon_\theta(z_\lambda;\lambda)-\epsilon\|_2^2
    \right],
    \label{eq:supp-kg-loss}
\end{equation}
where $w(\lambda)$ weights the log-SNR $\lambda=\log(\alpha_\lambda^2/\sigma_\lambda^2)$ and $p(\lambda)$ is its sampling density under the noise schedule. They show that this loss equals a negative ELBO,
\begin{equation}
    \mathcal L_w(x)
    =
    C
    -
    \mathbb E_{\lambda\sim\rho_w,\,\epsilon}
    \left[
        \mathrm{ELBO}_{\theta,\lambda}(z_\lambda)
    \right],
    \label{eq:supp-kg-elbo}
\end{equation}
where $C$ is independent of $\theta$ and $\rho_w$ is the nonnegative noise-level weighting induced by a monotone weight $w$. This condition holds for common schedules, including the linear schedule used by \algname{}. Eq.~\eqref{eq:supp-cfm-loss} is therefore, up to constants and the normalization $d_a$, a negative expected ELBO of the noise-augmented data.

The critic term uses the clipped proposal
\begin{equation}
    \bar\delta_\theta
    =
    \operatorname{clip}(\delta_\theta,-\alpha,+\alpha),
    \qquad
    \bar v_\theta=u_{\mathrm{buf}}+\bar\delta_\theta .
    \label{eq:supp-clipped-velocity}
\end{equation}
Since $\|\bar\delta_\theta\|_\infty\le\alpha$,
\begin{equation}
    \frac{1}{d_a}\|\bar v_\theta-u_{\mathrm{buf}}\|_2^2
    =
    \frac{1}{d_a}\|\bar\delta_\theta\|_2^2
    \le
    \alpha^2 .
    \label{eq:supp-clipped-cfm-bound}
\end{equation}
By the CFM--ELBO correspondence, the clipped proposal satisfies the local surrogate bound
\begin{equation}
    \mathbb E\left[
    \mathrm{ELBO}^{\mathrm{CFM}}_{\bar v_\theta}(z_\lambda)
    \right]
    \ge
    C-\alpha^2,
    \label{eq:supp-clipped-elbo-bound}
\end{equation}
with the same dimension-normalized scaling as Eq.~\eqref{eq:supp-cfm-loss}. The bound holds for the clipped proposal $\bar v_\theta$, not for the policy's own velocity $v_\theta$, which the flow-matching loss only keeps near it. Because $\bar v_\theta$ is built from $u_{\mathrm{buf}}$, which already encodes $a_{\mathrm{buf}}$, Eq.~\eqref{eq:supp-clipped-elbo-bound} bounds a surrogate rather than the likelihood of $a_{\mathrm{buf}}$ under the policy. A large CFM residual means that the noised replay action has a low ELBO surrogate under the actor, so clipping restricts value-guided updates to a trust region. The separate CFM loss is still applied to the unclipped velocity $v_\theta$: saturated coordinates receive no Q-gradient but are still pulled back toward $u_{\mathrm{buf}}$.

\paragraph{Action-space clamp.}
Each actor update draws one $(\epsilon, \tau)$ pair per replay sample, computes $x_\tau = (1-\tau)\epsilon + \tau a_{\mathrm{buf}}$, predicts $v_\theta(s, x_\tau, \tau)$, and forms the one-step reconstruction $\hat{x}_1 = x_\tau + (1-\tau)(u_{\mathrm{buf}} + \bar\delta_\theta)$ on which the Q term is evaluated. \algname{} clips the residual in \emph{velocity} space, $\bar\delta_\theta = \mathrm{clip}(\delta_\theta, -\alpha, \alpha)$, so the endpoint deviation $\hat{x}_1 - a_{\mathrm{buf}} = (1-\tau)\bar\delta_\theta$ shrinks linearly in $1-\tau$.
The ablation in Fig.~\ref{fig:clipping-compare} instead bounds the endpoint deviation directly in \emph{action} space. Enforcing $|\hat{x}_1 - a_{\mathrm{buf}}|\leq \alpha$ uniformly in $\tau$ rescales the bound on the policy-buffer gap by $1/(1-\tau)$, giving $\mathrm{clip}\bigl(\delta_\theta, -\tfrac{\alpha}{1-\tau}, \tfrac{\alpha}{1-\tau}\bigr)$. As $\tau\to1$ this bound grows without limit, so the same endpoint deviation corresponds to very different velocity magnitudes at different $\tau$. Optimizing the actor under this bound can therefore push $v_\theta$ toward sharply different magnitudes across $\tau$, producing the high-curvature flow field in Fig.~\ref{fig:flowfield-clamp-pair}.

\section{MuJoCo-v4 Settings}
\label{supp:mujoco}

All methods in Sec.~\ref{sec:exp:mujoco} share a CleanRL-based TD3 harness~\citep{huang2022cleanrl}, the flow actor and critic of Appendix~\ref{supp:qf3}, the Gymnasium MuJoCo-v4 tasks~\citep{towers2026gymnasium}, and the same evaluation protocol. Each run trains for $500$k environment steps and is evaluated every $10$k steps on $5$ deterministic episodes, with $3$ seeds per configuration. We tune each method's hyperparameters per task, selecting the configuration with the highest mean evaluation return over steps $450$k--$500$k, averaged over seeds; Table~\ref{tab:mujoco-hyperparameters} lists the grids and the selected settings. Curves show the mean $\pm$ population standard deviation across seeds.

\begin{table*}[t]
    \centering
    \footnotesize
    \begin{tabularx}{\linewidth}{@{}l
                                    >{\raggedright\arraybackslash}X
                                    >{\raggedright\arraybackslash}X@{}}
        \toprule
        Method & Searched grid & Selected setting (Hopper / Walker2d / Ant / Humanoid) \\
        \midrule
        QSM & $M_q \in \{25,50,120\}$ & $M_q = 25$ / $50$ / $50$ / $120$ \\
        \addlinespace
        FlowRL & $\lambda \in \{0.01,0.03,0.1,0.3,1\}$, expectile $\in \{0.7,0.8,0.9\}$ & $(\lambda,\text{expectile}) = (0.1,0.9)$ / $(1,0.8)$ / $(0.3,0.9)$ / $(0.03,0.8)$ \\
        \addlinespace
        DIPO & grad.-norm ratio $r \in \{0.03,0.08,0.1,0.3\}$, action-gradient steps $\in \{10,20,40\}$ & $(r,\text{steps}) = (0.3,20)$ / $(0.3,10)$ / $(0.03,40)$ / $(0.3,10)$ \\
        \addlinespace
        \algname{} & $\alpha\in\{0.5,1,2\}$, $\lambda_{\mathrm{cfm}}\in\{0.01,0.1,0.2,0.3,0.4,0.5\}$ & $(\alpha,\lambda_{\mathrm{cfm}}) = (1,0.2)$ / $(2,0.1)$ / $(1,0.5)$ / $(2,0.5)$ \\
        \bottomrule
    \end{tabularx}
    \caption{\textbf{MuJoCo-v4 method-specific settings.} Settings not listed are inherited from the shared CleanRL TD3 harness.}
    \label{tab:mujoco-hyperparameters}
\end{table*}

\paragraph{QSM.}
We implement Q-Score Matching~\citep{psenka2023qsm} following the authors' DDPM/JAX code. At noised replay actions, QSM regresses the velocity field onto the scaled critic action-gradient, $v_\theta \to M_q\,\nabla_a Q(s, a_\tau)$, without backpropagating through the sampler, and trains a twin critic with TD targets. The score-loss weight $M_q$ sets the magnitude of this target, and its best value differs across tasks (Fig.~\ref{fig:supp-qsm-hp-sweep}).

\paragraph{FlowRL and DIPO.}
We port FlowRL~\citep{lv2025flowbased} and DIPO~\citep{yang2023dipo} to this setup. FlowRL maximizes $Q$ with a Wasserstein-2 penalty toward high-value replay actions, which it identifies with an expectile value function; we tune the penalty weight $\lambda$ and the expectile. DIPO improves replay actions by gradient ascent on the critic and refits the actor to them; we tune its action-gradient norm ratio $r$ and the number of action-gradient steps. DIPO uses the flow actor in place of its original diffusion actor.

\paragraph{\algname{}.}
We tune \algname{}'s clip radius $\alpha$ and CFM weight $\lambda_{\mathrm{cfm}}$ over the values in Table~\ref{tab:mujoco-hyperparameters}. Fig.~\ref{fig:supp-qf3-velocity-sweep} shows the return over the last $50$k steps for each combination.

\IfFileExists{figures/qsm_hp_sensitivity.pdf}{%
\IfFileExists{figures/qf3_velocity_sensitivity.pdf}{%
\begin{figure*}[t]
    \centering
    \begin{subfigure}{0.8\linewidth}
        \centering
        \includegraphics[width=\linewidth]{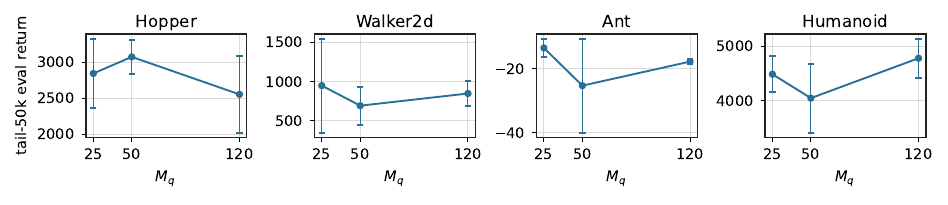}
        \caption{QSM score-loss weight $M_q$.}
        \label{fig:supp-qsm-hp-sweep}
    \end{subfigure}
    \par\vspace{0.7em}
    \begin{subfigure}{0.8\linewidth}
        \centering
        \includegraphics[width=\linewidth]{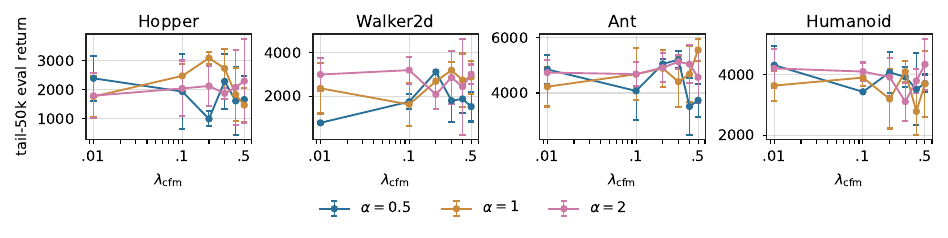}
        \caption{\algname{} velocity clip radius $\alpha$ and CFM weight $\lambda_{\mathrm{cfm}}$.}
        \label{fig:supp-qf3-velocity-sweep}
    \end{subfigure}
    \caption{\textbf{MuJoCo-v4 hyperparameter sensitivity.} Tail-50k evaluation return; markers show across-seed means and error bars $\pm 1\sigma$ over $3$ seeds.}
    \label{fig:supp-hp-sweeps}
\end{figure*}
}{}%
}{}

\section{Holosoma Humanoid Locomotion Settings}
\label{supp:holosoma}

This section gives the simulator, task, and baseline configuration for the velocity- and motion-tracking experiments in Sec.~\ref{sec:exp:locomotion}.

\paragraph{Simulator and training stack.}
All locomotion experiments run in the open-source Holosoma humanoid whole-body RL framework~\citep{amazonfar2025holosoma}, which provides (i) a GPU-parallel humanoid simulator, (ii) velocity-tracking and single-motion-tracking tasks for the Unitree G1, and (iii) a tuned FastTD3~\citep{seo2025fasttd3} baseline. We use Holosoma's default observation processing, action space (29-DoF joint targets fed to a low-level PD controller), domain randomization, and termination conditions. \algname{} replaces the actor inside the same FastTD3 training loop, sharing the replay buffer, distributional twin-Q critic, target-update schedule, and parallel-environment rollouts. The FastTD3 baseline in Fig.~\ref{fig:locomotion-curves} uses Holosoma's tuned configuration unchanged. Velocity-tracking runs train on a single NVIDIA A10G GPU and motion-tracking runs on a single NVIDIA L40S GPU.
Both tasks use the Unitree G1 (29 actuated DoFs) as released in Holosoma. The same URDF, joint limits, and PD gains are used in training and in zero-shot deployment; the policy outputs joint targets at the simulator's control rate, which the on-robot low-level controller tracks.

\paragraph{Tasks and command distribution.}
We use Holosoma's two reference tasks unchanged. Velocity tracking commands a planar base velocity $(v_x, v_y, \omega_z)$, resampled per episode from Holosoma's default command distribution. The y-axis of Fig.~\ref{fig:splash-vel-curves} is the per-step \texttt{raw\_lin\_vel} reward, Holosoma's unweighted exponentiated linear-velocity tracking term. Single-motion tracking uses the LAFAN \texttt{dance1\_subject2} clip~\citep{harvey2020lafan} as its reference; the y-axis of Fig.~\ref{fig:splash-mot-curves} is the mean episode length, capped at $500$ steps.

\paragraph{\algname{} hyperparameters.}
\algname{} and FastTD3 share the actor and critic learning rates, target smoothing $\rho$, batch size, replay capacity, parallel environment count, gradient steps per environment step, and exploration noise, and on velocity tracking both use Holosoma's left--right symmetry augmentation. Network widths and the distributional critic are the ``Locomotion'' column of Table~\ref{tab:qf3-arch}. \algname{} uses $\alpha=2$ and $\lambda_{\mathrm{cfm}}=0.3$ on velocity tracking and $\alpha=0.5$ and $\lambda_{\mathrm{cfm}}=0.01$ on motion tracking, and integrates the flow with $5$ Euler steps.

\paragraph{FPO++ baseline.}
The FPO++~\citep{yi2026fpopp} curves in Fig.~\ref{fig:locomotion-curves} are produced with the authors' open-source code and its released configuration, unchanged. The wall-clock comparison runs all three methods on the same GPU type for each task (A10G for velocity tracking, L40S for motion tracking).

\paragraph{FlowRL baseline.}
FlowRL~\citep{lv2025flowbased} uses the same FastTD3 recipe, replacing the actor and its loss with FlowRL's. Its velocity-tracking runs train on a single NVIDIA L4 GPU rather than an A10G, and its motion-tracking runs on a single NVIDIA L40S, so Fig.~\ref{fig:locomotion-curves} omits FlowRL from the velocity wall-clock panel.

\section{ABC Sim Manipulation Fine-tuning Settings}
\label{supp:manipulation}

This section gives the policy, training, baseline, and evaluation configuration for the manipulation experiments in Sec.~\ref{sec:exp:manipulation}. Both tasks use 3 seeds per method.

\paragraph{Base policy.}
We fine-tune the vision-conditioned ABC-VLA policy~\citep{abc2026}. A frozen Gemma+SigLIP VLM (4.3B parameters) conditions an 8-block, 44M-parameter DiT action head. The head integrates a 5-step Euler ODE from Gaussian noise to a 30-step action chunk, of which the first 15 steps are executed. The dashed base-policy level in Fig.~\ref{fig:manipulation-finetuning} pools the step-0 evaluations of all methods.

\paragraph{Trainable parameters.}
\algname{} trains rank-4 LoRA adapters on the attention and MLP projections of the DiT head, 229k trainable parameters in total; the VLM and base head stay frozen. Adapters start at zero, so training begins exactly at the base policy. We omit the q/k/v adapters of the first four blocks.

\paragraph{Actor update.}
We apply the \algname{} objective of Sec.~\ref{sec:algorithm} to replay chunks: noise the stored chunk to a random $\tau$, query one velocity, clip the policy-buffer gap to $\pm\alpha$ ($\alpha=0.5$ on \emph{Bottles}, $0.1$ on \emph{Dishrack}), and score the one-step denoised chunk with the critic. Two anchors regularize the update: $\lambda_{\mathrm{cfm}}=0.01$ toward the replay chunk and $\lambda_{\mathrm{base}}=1.0$ toward the base head's velocity with adapters switched off. The unclamped ablation sets $\alpha\to\infty$ with the same anchors.

\paragraph{Training loop.}
Each iteration, 16 parallel worlds each collect one episode with the Polyak-averaged target actor ($\tau_{\mathrm{polyak}}=0.05$), followed by 1600 critic updates (batch 64, learning rate $3\times10^{-4}$) and 200 actor updates (batch 256, learning rate $4.5\times10^{-5}$). The replay buffer holds 50k transitions and is pre-filled with base-policy rollouts. No action noise is injected; exploration comes from the flow latent and the lag between the target and online actors.

\paragraph{Residual RL baseline.}
The flow policy is frozen, and a ResFiT-style MLP (2.8M parameters) reads the cached VLM condition, robot state, and base action and outputs a $\tanh$-bounded correction (scale $0.03$ on \emph{Bottles}, $0.01$ on \emph{Dishrack}) added to the executed action. It is trained with TD3 on the same critic, replay buffer, simulator, and evaluation protocol, and its zero-initialized output layer makes it start at the base policy.

\paragraph{Tasks and evaluation.}
An episode succeeds when every target object has been placed at least once, and ends on success or at the timeout. Rewards are per control step and discounted within the action chunk before entering the replay buffer.
\emph{Bottles}: five bottles, re-randomized every episode; reward $+1$ the first time each bottle enters the bin; 1000-step timeout; 50-episode evaluation on fresh layouts after every rollout; runs stopped at 400k environment steps.
\emph{Dishrack}: three plates in one fixed layout; reward $+1/3$ per newly racked plate and $+1$ on success; 3300-step timeout; 50-episode evaluation every 20k environment steps.
Success rate is the fraction of successful episodes. Mean episode length is averaged over successful episodes only (timeouts excluded), so it measures how quickly the policy finishes when it succeeds.

Table~\ref{tab:abc-settings} lists the simulation and rollout settings of both tasks.

\begin{table}[h]
    \centering
    \footnotesize
    \setlength{\tabcolsep}{4pt}
    \begin{tabular}{lcc}
        \toprule
         & \emph{Bottles} & \emph{Dishrack} \\
        \midrule
        Control rate & \multicolumn{2}{c}{$29.4$\,Hz ($500/17$)} \\
        Steps executed per policy call & \multicolumn{2}{c}{$15$ (${\approx}2$ calls/s)} \\
        Parallel worlds, training & $16$ & $16$ \\
        Warm-up worlds $\times$ episodes & $80\times4$ & $64\times2$ \\
        Eval.\ worlds $\times$ batches & $50\times1$ & $10\times5$ \\
        Max.\ episode length (steps) & $1000$ ($34$\,s) & $3300$ ($112$\,s) \\
        \bottomrule
    \end{tabular}
    \caption{\textbf{ABC Sim per-task settings.} Each evaluation runs $50$ episodes. Warm-up rollouts come from the base policy and pre-fill the replay buffer.}
    \label{tab:abc-settings}
\end{table}

\section{Robomimic Fine-tuning Settings}
\label{supp:robomimic}

This section gives the task, policy, and training configuration for the Robomimic experiments in Sec.~\ref{sec:exp:robomimic}. All methods run in the released OGPO codebase~\citep{patil2026ogpo} with its Robomimic settings; each task uses 3 seeds per method.

\paragraph{Tasks.}
\emph{Square} (single-arm peg insertion), \emph{Tool Hang} (single-arm multi-step tool insertion), and \emph{Transport} (bimanual hand-over) use state observations, and policies output end-effector pose deltas and gripper commands at $20$\,Hz. Every task has a sparse reward of $-1$ per step until success, and an episode ends on success or at the time limit. Per-task settings are listed in Table~\ref{tab:robomimic-settings}.

\begin{table}[h]
    \centering
    \footnotesize
    \setlength{\tabcolsep}{4pt}
    \begin{tabular}{lccc}
        \toprule
         & \emph{Square} & \emph{Tool Hang} & \emph{Transport} \\
        \midrule
        State / action dim. & 14 / 7 & 14 / 7 & 28 / 14 \\
        Episode limit & 400 & 1000 & 800 \\
        Demonstrations & MH & PH & MH \\
        BC stopped at ${\le}50\%$ & yes & yes & yes \\
        Action chunk $h$ & 4 & 8 & 8 \\
        Discount $\gamma$ & 0.99 & 0.999 & 0.999 \\
        Online steps & 1.5M & 2M & 2M \\
        \algname{} clip $\alpha$ & 0.12 & 0.12 & 0.02 \\
        \bottomrule
    \end{tabular}
    \caption{\textbf{Robomimic per-task settings.} Task and training settings follow OGPO~\citep{patil2026ogpo}. MH/PH: multi-/proficient-human demonstrations.}
    \label{tab:robomimic-settings}
\end{table}

\paragraph{Base policy and warm start.}
The base policy is an MLP flow policy integrated with $10$ flow steps, pretrained with behavior cloning on the demonstrations in Table~\ref{tab:robomimic-settings}. All methods start from the same checkpoint. No demonstrations are used during online RL; the replay buffer is seeded with base-policy rollouts before training starts.
Under the evaluation protocol below, the base checkpoint succeeds in $53\%$, $48\%$, and $48\%$ of episodes on \emph{Square}, \emph{Tool Hang}, and \emph{Transport} ($27\%$, $20\%$, and $14\%$ with a single deterministic sample).

\paragraph{Shared critic and training loop.}
All full fine-tuning methods share OGPO's critic: an ensemble of $10$ MLPs ($4\times512$ hidden units; $5\times512$ on \emph{Transport}), trained by TD regression with the ensemble mean as the target and Polyak averaging ($\tau=0.05$), with batch size $256$. A single (non-vectorized) environment collects training data, and each environment step is followed by one critic and one actor update.

\paragraph{\algname{} and \algname{}+.}
\algname{} replaces OGPO's actor update with the clipped one-step objective of Sec.~\ref{sec:algorithm}; its $Q$ term takes the minimum over all $10$ critic heads. Before the first actor update, the critic is trained alone on base-policy rollouts for $20$k steps on \emph{Square} and \emph{Tool Hang} and $100$k steps on \emph{Transport}. \algname{}+ adds OGPO's success-buffer term: a conditional flow-matching loss on actions from successful episodes in the replay buffer.
Both add the base-policy anchor of Sec.~\ref{sec:exp:finetuning} with $\lambda_{\mathrm{base}}=12.5$.
\algname{}+ weights the success-buffer loss by $100$, larger than OGPO's $1.0$, because it is balanced against the $-Q$ term, whose magnitude is on the scale of the returns.

\paragraph{OGPO.}
We run OGPO+CA with the authors' Robomimic hyperparameters: PPO-style updates over $32$ sampled denoising trajectories per state with the critic as the terminal reward, the success-buffer loss (weight $1.0$), and conservative advantages, which take the minimum advantage over the critic ensemble when all members agree it is positive, the maximum when all agree it is negative, and zero otherwise.

\paragraph{DSRL and EXPO.}
DSRL~\citep{wagenmaker2025steering} and EXPO~\citep{dong2025expo} use the implementations and Robomimic settings of the OGPO codebase and start from the same checkpoint. Neither updates the base policy with RL: DSRL keeps the checkpoint frozen and learns a policy over its input noise, and EXPO learns a separate policy that edits the base policy's actions.

\begin{figure}[t]
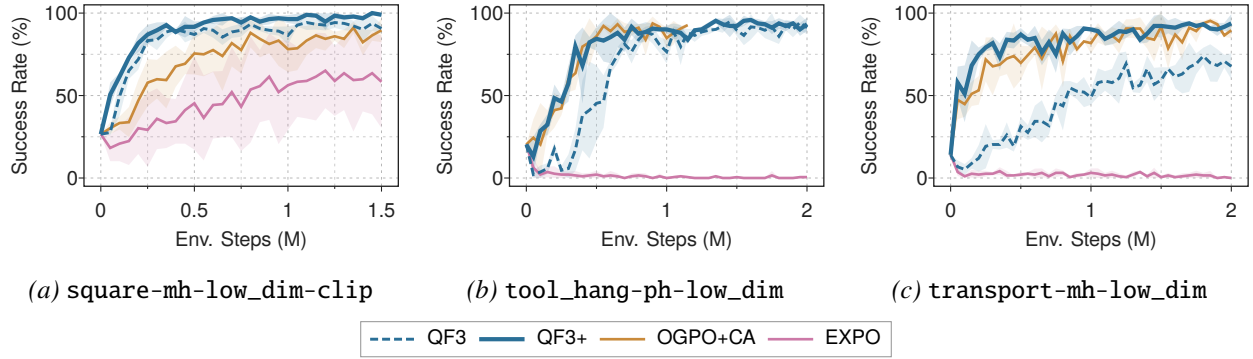

    \centering
    \flopRobomimicOde{figures/data/robomimic/tidy_ode}
    \caption{\textbf{Robomimic fine-tuning, deterministic evaluation.} The runs of Fig.~\ref{fig:robomimic}, evaluated with one ODE sample of the flow policy instead of the policy each method acts with: no SDE noise and no best-of-$N$ selection by the critic. Step $0$ is the shared checkpoint ($27\%$, $20\%$, and $14\%$). 3 seeds, mean $\pm$ std.}
    \label{fig:robomimic-ode}
\end{figure}

\paragraph{Evaluation.}
Every $50$k online steps we run $64$ evaluation episodes ($32$ parallel environments, $2$ episodes each). \algname{}, \algname{}+, and OGPO+CA act with OGPO's best-of-$N$ sampler, in training rollouts and in evaluation: they draw $8$ candidate actions with the SDE sampler and execute the one the critic scores highest. EXPO is evaluated with its on-the-fly policy, which also selects among base and edited samples with the critic, and DSRL samples from its noise policy. Success rate is the fraction of successful episodes.
Fig.~\ref{fig:robomimic-ode} reports the same runs evaluated with a single deterministic (ODE) sample of the flow policy, without the critic.

\section{ABC Sim Speedup Diagnostics}
\label{supp:manipulation-diagnostics}

We analyze the gains of Sec.~\ref{sec:exp:manipulation}, and for each task, we run the base policy and a final \algname{} checkpoint on the same 50 evaluation seeds.

\subsection{Load Plates in Dishrack}

\algname{} raises success from $48\%$ to $72\%$ and shortens the mean episode from $90$ to $63$\,s. About three quarters of this reduction comes from the 15 seeds that the base policy fails and \algname{} solves; 21 seeds succeed under both, 3 only under the base policy, and 11 under neither.

\begin{figure*}[p]
    \centering
    \includegraphics{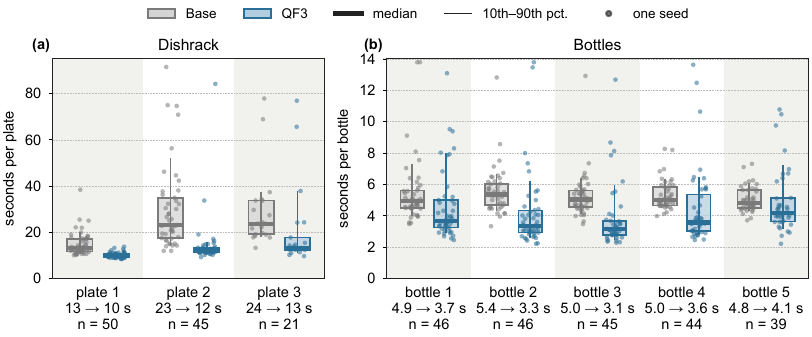}
    \caption{\textbf{Time per object}, from one object being counted to the next, on the seeds where both policies complete that object. Labels give the median for the base policy $\rightarrow$ \algname{} and the number of seeds. \emph{Bottles} uses a separate recorded evaluation and omits the 4 layouts with a bottle counted at the first step; points above $14$\,s are drawn at the top edge.}
    \label{fig:supp-manip-times}
    \par\vspace{0.5em}
    \includegraphics{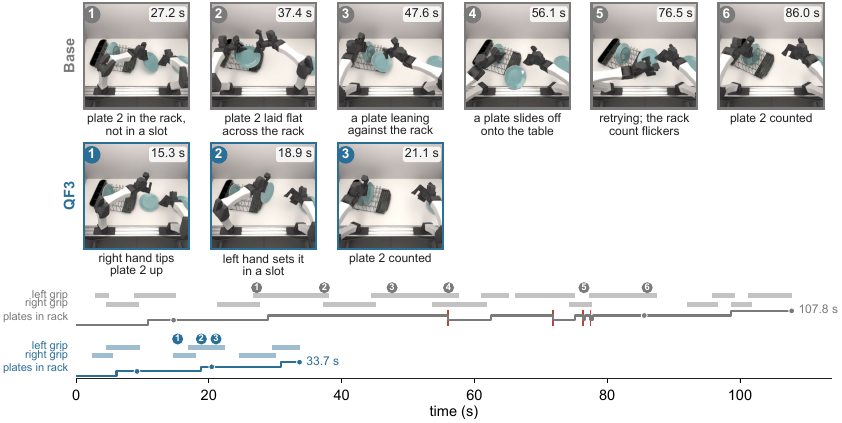}
    \caption{\textbf{One \emph{Dishrack} seed.} The base policy spends $71$\,s on the second plate and \algname{} $11$\,s. Stills are Blender renders of the recorded states, numbered as on the timeline, which shows gripper closures, plates in the rack (red ticks: a plate leaves), and evaluation counts (dots).}
    \label{fig:supp-dishrack-filmstrip}
    \par\vspace{0.5em}
    \includegraphics{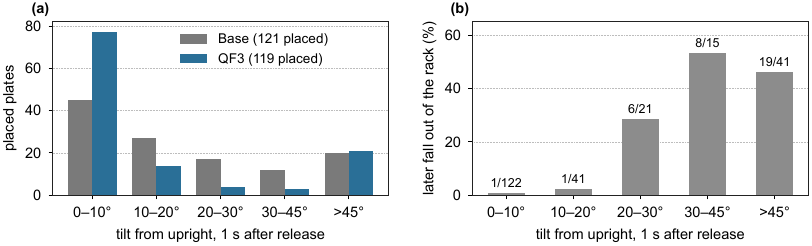}
    \caption{\textbf{Plate tilt on \emph{Dishrack}.} (a) Tilt from upright $1$\,s after release, per policy. (b) Share of placed plates in each tilt band that later leave the rack, both policies pooled.}
    \label{fig:supp-manip-tilt}
\end{figure*}

\algname{} converges on strong behaviors that the base policy already produces. Here, we find \algname{}'s biggest gains are not from \textit{faster} actions: about four fifths of the per-plate time saved is in manipulation (from the first grasp that moves a plate until it is counted) rather than reach, and on the first plate \algname{} even reaches more slowly ($4.8$ vs.\ $3.0$\,s) and carries the plate into the rack no faster (median $5.1$--$5.2$\,s for both). Instead, it decreases the number of handoffs. On the first plate, the base policy usually grasps with the left hand, lifts the plate only a few centimeters, regrasps with the right, and takes it back with the left (41 of 50 episodes); \algname{} hands it directly from the right hand to the left (49 of 50), a handover the base policy already uses on 4 seeds at nearly the same speed (median $10.6$ vs.\ $9.9$\,s).

For the dish insertion: on later plates, the base policy often pushes a plate into the rack without seating it in a slot, and then keeps adjusting it, lays it across the wires, or hands it back and retries, sometimes for over a minute (Fig.~\ref{fig:supp-dishrack-filmstrip}). \algname{}'s carry-and-insert is short and consistent ($7.3$ vs.\ $12.0$\,s when successful).
In the base model, a plate left more than $20^\circ$ from upright often slides back out (Fig.~\ref{fig:supp-manip-tilt}b), typically $14$\,s, or about 27 policy calls, after its release and usually with neither gripper within $16$\,cm (contact with another part of the arm cannot be ruled out). \algname{} learns to seat plates nearly upright (Fig.~\ref{fig:supp-manip-tilt}a) and loses $8\%$ of the plates it places, against $21\%$ for the base policy.

\subsection{Throw Bottles in Bin}

\algname{} raises success from $86\%$ to $96\%$ and shortens the mean episode from $26.9$ to $21.1$\,s. Unlike on \emph{Dishrack}, about three quarters of this reduction comes from seeds that both policies solve, and the improvement comes from \algname{} speeding up episodes. The typical bottle takes $1$--$2$\,s less (Fig.~\ref{fig:supp-manip-times}b), and is held shorter ($1.8$ vs.\ $2.5$\,s) and grasps sooner.

\section{Text-to-Image Fine-tuning}
\label{supp:t2i}

We apply the \algname{} actor update to text-to-image fine-tuning, as a demonstration beyond robotics.

\paragraph{Setup.}
We fine-tune Stable Diffusion 3.5 Medium~\citep{esser2024sd3} to maximize PickScore~\citep{kirstain2023pickapic}, keeping the sampler, reward model, and LoRA setup of the FDFO codebase~\citep{mcallister2026fdfo}. We treat text-to-image generation as a one-step RL problem: the prompt $c$ is the observation, the final image latent is the action, and the reward model $R$ replaces the learned critic. For a replayed latent $a_{\mathrm{buf}}$, with $x_\tau$, $u_{\mathrm{buf}}$, and $\delta_\theta$ as in Eq.~\eqref{eq:supp-cfm-path} and $c$ in place of $s$, the actor minimizes
\begin{equation}
    -R\bigl(D(\hat{x}_1),\,c\bigr)
    + \lambda_{\mathrm{cfm}}\,\|\delta_\theta\|^2
    + \lambda_{\mathrm{base}}\,\|v_\theta - v_{\mathrm{base}}\|^2,
    \qquad
    \hat{x}_1 = a_{\mathrm{buf}} + (1-\tau)\operatorname{clip}(\delta_\theta, -\alpha, \alpha),
    \label{eq:supp-t2i-loss}
\end{equation}
where $D$ is the VAE decoder and $v_{\mathrm{base}}$ is the network with the LoRA adapter disabled. The reward is evaluated at the clipped one-step endpoint, with its gradient flowing through $D$ and PickScore's image encoder, while the flow-matching term acts on the unclipped residual.

\paragraph{Training.}
We train a rank-32 LoRA adapter (scale $64$) on the transformer's attention projections with AdamW (learning rate $3\times10^{-5}$, weight decay $10^{-4}$, gradient-norm clip $1.0$), following FDFO. Each epoch samples $128$ images at $512\times512$ ($40$ sampler steps, no classifier-free guidance) for fresh prompts from PickScore's $25{,}432$ training prompts, adds their latents to a $1024$-entry first-in-first-out replay buffer, and takes $16$ optimizer steps on $64$ replay draws each. We use $\alpha=0.225$, about half the mean $|\delta_\theta|$ (clipping $63$--$67\%$ of elements), $\lambda_{\mathrm{cfm}}=0.01$, and $\lambda_{\mathrm{base}}=1$, and train for $24$ epochs on one NVIDIA H100.

We test on three settings: a replay buffer of size $128$, refreshed every epoch; a replay buffer of length $1024$, sampled uniformly; and a buffer of length $1024$ but where draws favor recent entries, with weight $0.5^{k/128}$ for the $k$-th newest, such that about half come from the current epoch. We also run the uniform $1024$ configuration with the clip removed, so that $\hat{x}_1$ uses the unclipped $\delta_\theta$.

\begin{figure*}[p]
    \centering
    \begin{subfigure}[b]{0.365\linewidth}
        \centering
        \includegraphics[width=\linewidth]{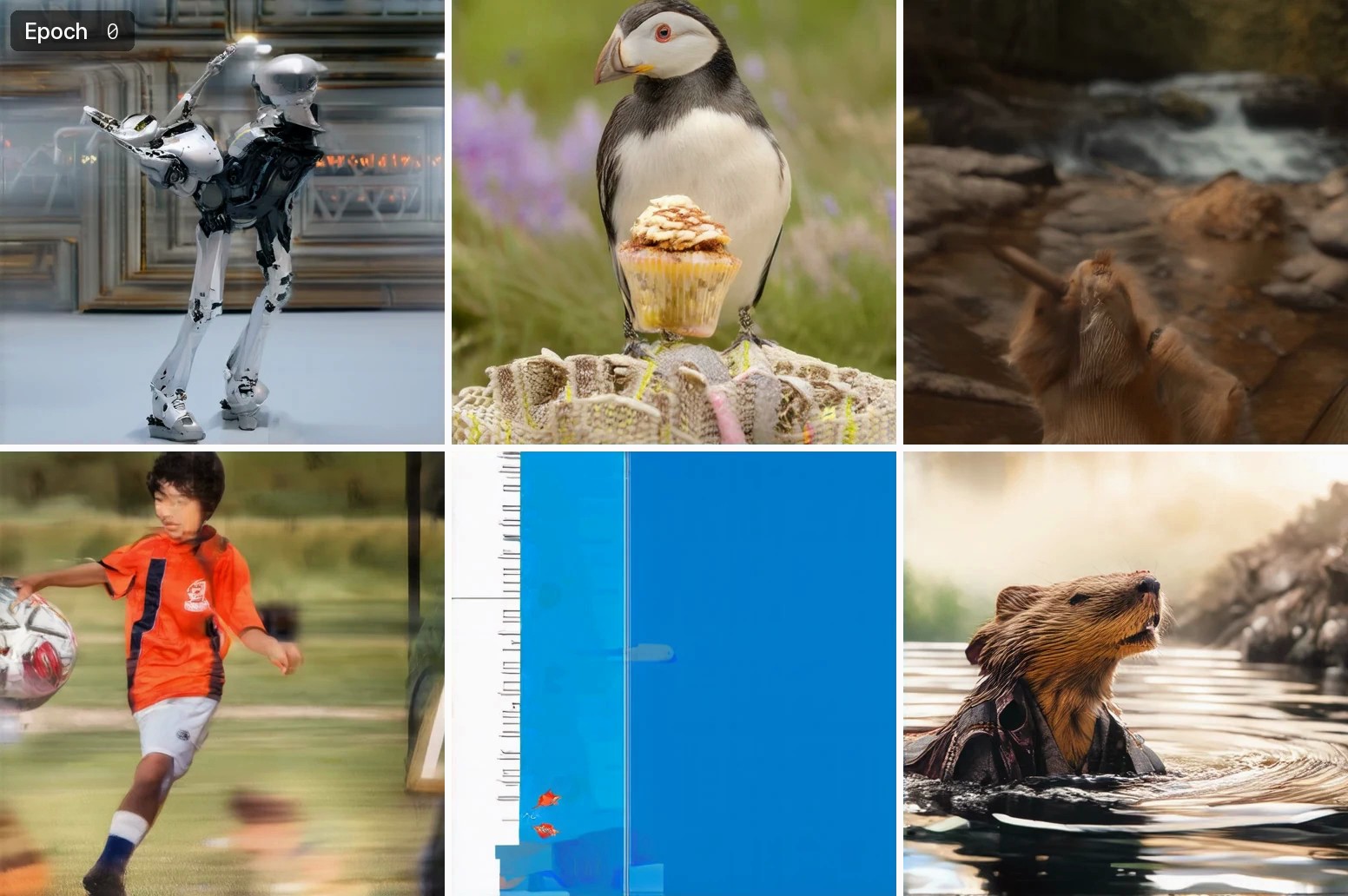}
        \caption{Base model.}
    \end{subfigure}\hfill
    \begin{subfigure}[b]{0.365\linewidth}
        \centering
        \includegraphics[width=\linewidth]{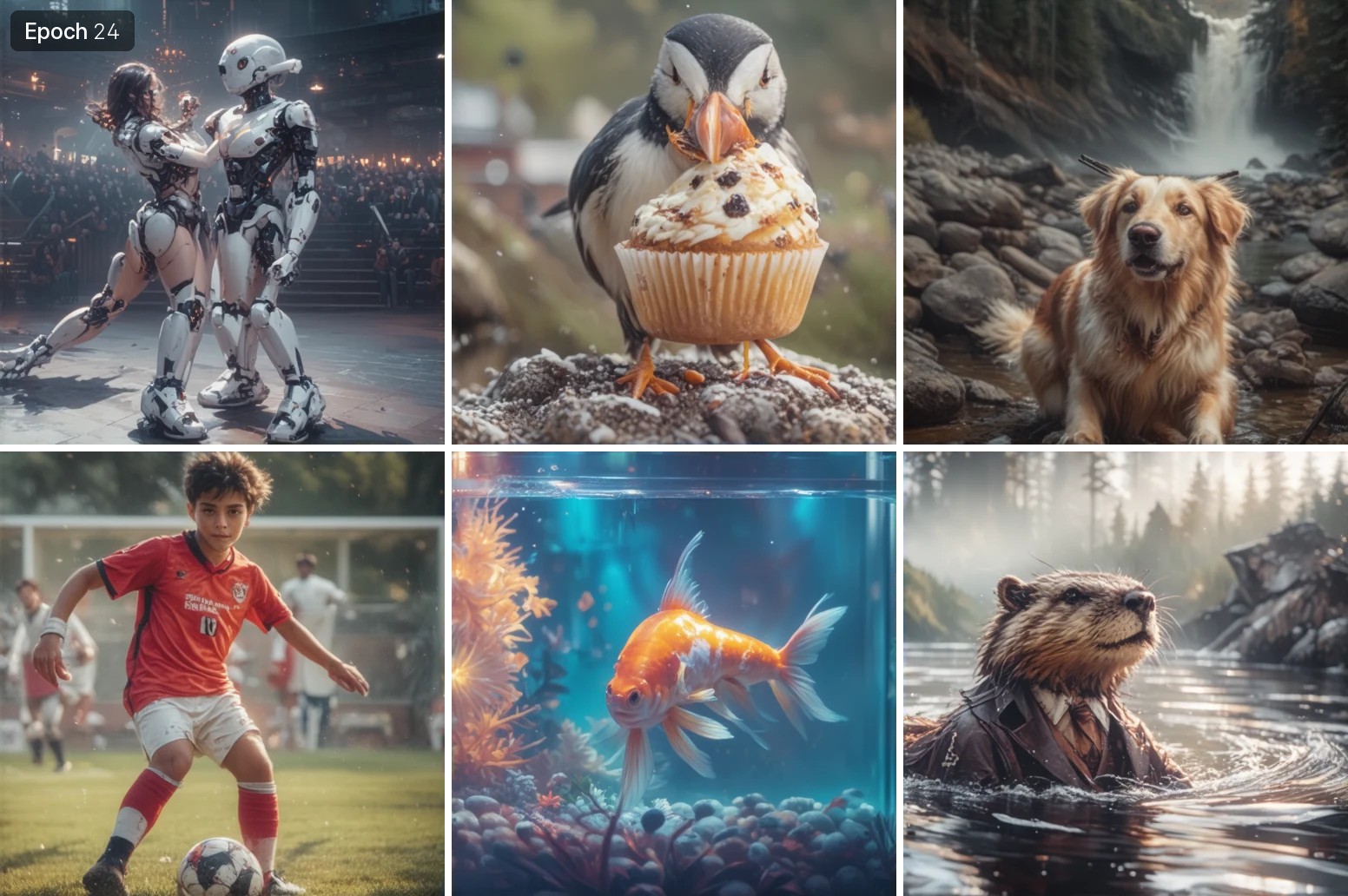}
        \caption{After $24$ epochs of \algname{}.}
    \end{subfigure}\hfill
    \begin{subfigure}[b]{0.24\linewidth}
        \centering
        \begin{tikzpicture}
        \begin{axis}[flop curves,
            width=0.74\linewidth, height=0.6\linewidth,
            xmin=0, xmax=23, xtick={0,10,20}, xticklabels={0,10,20},
            ymin=19.5, ymax=23, ytick={20,21,22,23},
            xlabel={Epoch}, ylabel={Train PickScore}]
          \addplot[color=flopAccentA, line width=1.2pt, mark=*, mark size=0.8pt]
            table[col sep=comma, x=epoch, y=pickscore, comment chars={\#}]
            {figures/data/t2i/pickscore_qf3_hl128.csv};
        \end{axis}
        \end{tikzpicture}
        \caption{Training reward.}
    \end{subfigure}
    \caption{\textbf{Fine-tuning Stable Diffusion 3.5 Medium on PickScore.} (a, b) Six fixed prompts (Table~\ref{tab:t2i-prompts}) and seeds before and after fine-tuning. (c) Training reward: mean PickScore of each epoch's training samples.}
    \label{fig:supp-t2i}
    \par\vspace{1em}
    \setcounter{subfigure}{0}
    \begin{subfigure}{\linewidth}
        \centering
        \includegraphics[width=\linewidth]{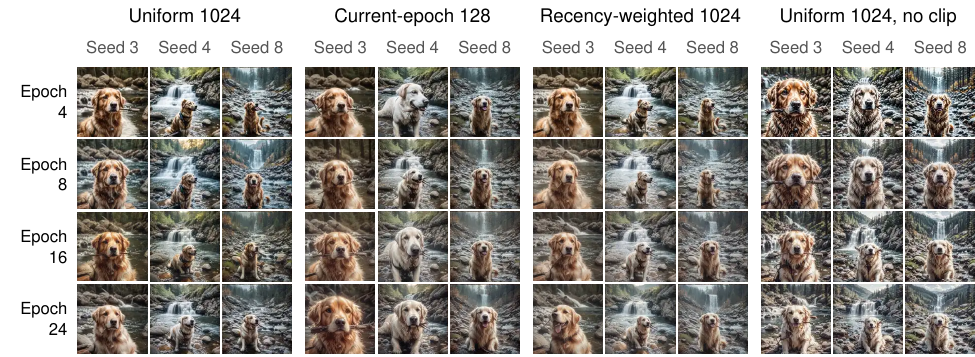}
        \caption{Golden retriever.}
    \end{subfigure}
    \par\vspace{0.35em}
    \begin{subfigure}{\linewidth}
        \centering
        \includegraphics[width=\linewidth]{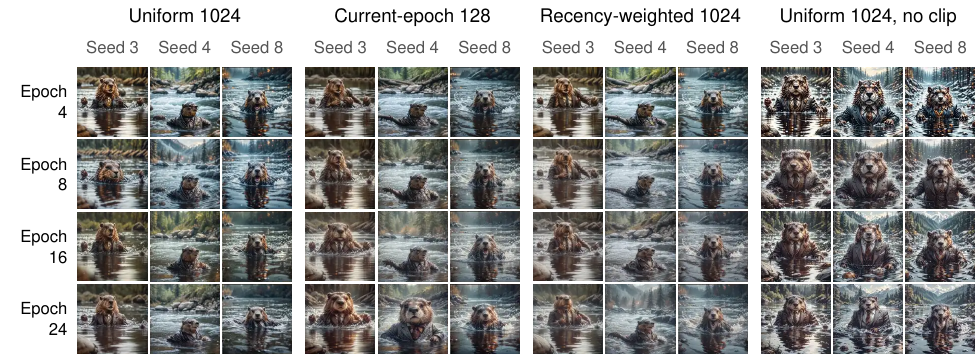}
        \caption{Beaver in a suit.}
    \end{subfigure}
    \caption{\textbf{Qualitative progression under three replay configurations and without the clip.} Columns compare methods; the last is uniform replay ($1024$) with the clip removed. Rows show epochs $4$, $8$, $16$, and $24$. Each method uses the same three seeds.}
    \label{fig:supp-t2i-progression}
\end{figure*}

\paragraph{Results.}
We report the results in Fig.~\ref{fig:supp-t2i}, and the results show the mean PickScore of the training samples rises from $19.72$ to $22.57$; Table~\ref{tab:t2i-prompts} lists the prompts shown in Fig.~\ref{fig:supp-t2i}.
Figure~\ref{fig:supp-t2i-progression} follows the same illustrative seeds ($3$, $4$, $8$) at epochs $4$, $8$, $16$, and $24$ for the golden-retriever and beaver prompts. All samples use $40$ sampler steps, guidance $1.0$, and $512\times512$ generation. These examples illustrate qualitative differences across methods.

Despite similar training PickScore, the replay configurations appear to follow different qualitative trajectories (Fig.~\ref{fig:supp-t2i-progression}). In the displayed probes, current-epoch replay ($128$) tends toward more uniform, close-up compositions, reducing variation in subject scale and framing. Uniform replay ($1024$) retains more varied compositions but exhibits larger fluctuations in color, contrast, and lighting across epochs. Recency-weighted replay ($1024$) maintains a more consistent visual appearance while retaining variation in framing.

\begin{table}[b]
    \centering
    \footnotesize
    \begin{tabular}{ll}
        \toprule
        Position & Prompt \\
        \midrule
        Top left & ``a robot doing the tango'' \\
        Top middle & ``a puffin eating a muffin'' \\
        Top right & ``golden retriever with a stick in its mouth on a hike to a waterfall'' \\
        Bottom left & ``a boy kicking a soccer ball'' \\
        Bottom middle & ``goldfish swimming in a fishtank'' \\
        Bottom right & ``beaver wearing a suit swimming in a river'' \\
        \bottomrule
    \end{tabular}
    \caption{\textbf{Prompts in Fig.~\ref{fig:supp-t2i}}, by position in each grid.}
    \label{tab:t2i-prompts}
\end{table}

\end{document}